\documentclass[withindex,glossary]{cam-thesis}
\pdfoutput=1
\usepackage[numbers]{natbib}
\usepackage{algorithm}
\usepackage[noend]{algpseudocode}
\usepackage{amsmath}
\usepackage{listings}
\usepackage{csquotes}
\usepackage{amssymb}
\usepackage{tabulary}
\usepackage{siunitx}
\usepackage{gensymb}
\usepackage{multirow}
\usepackage{booktabs}
\usepackage{float}
\usepackage{graphicx}
\usepackage{makecell}
\usepackage{wrapfig}
\usepackage{tikzpagenodes}
\usepackage{subfig}
\usepackage{xcolor}
\definecolor{light-gray}{gray}{0.95}

\title{Vision based body gesture meta features for Affective Computing}

\author{Indigo Jay Dennis Orton}

\college{Hughes Hall}

\date{June, 2019}

\subjectline{Computer Science}

\abstract{%

\subsubsection*{Title: Vision based body gesture meta features for Affective Computing}

\noindent
Early detection of psychological distress is key to effective treatment.
Automatic detection of distress, such as depression, is an active area of research.

Current approaches utilise vocal, facial, and bodily modalities.
Of these, the bodily modality is the least investigated, partially due to the difficulty in
extracting bodily representations from videos, and partially due to the lack of viable datasets.
Existing body modality approaches use automatic categorization of expressions to represent body
language as a series of specific expressions, much like words within natural language.

In this dissertation I present a new type of feature, within the body modality, that represents meta
information of gestures, such as speed, and use it to predict a non-clinical depression label.
This differs to existing work by representing overall behaviour as a small set of aggregated meta
features derived from a person's movement.
In my method I extract pose estimation from videos, detect gestures within body parts, extract meta information
from individual gestures, and finally aggregate these features to generate a small feature vector
for use in prediction tasks.

No existing, publicly available, dataset for distress analysis contains source videos of participants
or pose data extracted from the source videos.
As such, I introduce a new dataset of 65 video recordings of interviews with self-evaluated distress,
personality, and demographic labels.
This dataset enables the development of features utilising the whole body in distress detection tasks.

I evaluate my newly introduced meta-features for predicting depression, anxiety, perceived stress,
somatic stress, five standard personality measures, and gender.
A linear regression based classifier using these features achieves a 82.70\% F1 score for predicting depression within my
novel dataset.

My results suggest this feature type has value for distress prediction and, more broadly, has
potential within the affective computing domain, as they appear to be useful aggregate
representations of human behaviour.

\newpage
Word count: 14,586
}

\acknowledgements{%
First and foremost I thank my supervisor Dr. Marwa Mahmoud, her guidance has been invaluable and
her collaboration in the development of the dataset has been integral.
Thank you to my peers who have contributed immensely to my learning through countless rambling and
broad ranging discussions.
Their breadth of interests and expertise is always engaging.
Finally, thank you to my family for their endless support and encouragement.
}

\newglossaryentry{CNN}{
name=CNN,
description={Convolutional Neural Network}
}

\begin{document}

  \frontmatter{}

\chapter{Introduction}\label{ch:introduction}

Mental health is an increasingly prominent area of discussion and research in society, it is
a main contributor to overall disease burden globally~\cite{mentalhealthuk}.
Awareness of its effects and symptoms, destigmatization of its results, consideration of its causes,
and approaches to its treatment and diagnosis, in the case of illness, are all rapidly evolving.
According to the UK Mental Health Foundation~\cite{mentalhealthuk}, depression is the \enquote{predominant
mental health problem worldwide}.
Within the UK in 2014, 19.7\% of people over the age of 16 showed symptoms of depression or anxiety.
Early detection is paramount to long term health, 10\% of children aged 5--16 years have a clinically
diagnosable mental health issue, of children who have experienced mental health problems 70\% do not
have their issues detected or addressed early enough~\cite{mentalhealthuk}.

\subsection*{Automatic Distress Detection - Why and How}
Automatic distress detection enables large scale early screening.
Early screening can be used to enable mitigation of distress at an earlier stage that it might
otherwise be identified and also prioritization of services.
For example, \textit{Crisis Text Line}~\cite{CTL_HOME} (CTL) provides a free mental health counseling
service via text messages.
The communication medium means consumers can message without a counselor being exclusively available.
To better support those consumers most in need (e.g.\ those most at risk of suicide), CTL analyses
text messages in real time to triage consumers for counselor attention~\cite{CTL_BLOG}.

While the CTL example is a specific use case with a single modality, text, the broader area of
distress detection uses many modalities including facial, eye, head, vocal, bodily, and speech.
Existing work has investigated usage of psychology coding systems such as FACS
levels of activity in eye and head movement~\cite{Alghowinem:2015cs,Syed:2017jb,Dibeklioglu:2015cf},
orientation of eye gaze and head~\cite{Scherer:2014is}, fundamental frequencies and biomarkers
of vocal recordings~\cite{Ozdas:do,Alghowinem:2016jq}, automatic categorization of body
expressions~\cite{Joshi:uf,Joshi:2013bp,Song:2013ik}, and natural language artifacts
from speech~\cite{Dang:2017fd}.
Moreover, many methods incorporate multiple modalities and achieve significantly better results.

One of the difficulties in this area is the lack of available and relevant datasets.
Distress focused datasets are not readily shared given the sensitive nature of the subject.
Those that are shared rarely include source data, but rather provide processed outputs, again due
to the sensitive of the subject.
Finally, each modality has different source data requirements and datasets are generally designed
with a specific method in mind.
All of this creates a barrier to the development of novel features and thus research within this
area is concentrated on the modalities available within the few public datasets.

\subsection*{Body Modality - A Research Gap}
The body modality includes the whole body from the neck down, sometimes the head is also included.
It is the least researched of the modalities.
This is, partially, due to the dataset barrier.
No public distress dataset, that I am aware of, provides body modality data, or the source video
recordings to extract such data.
The majority of relevant body modality based automatic distress detection work is based on private clinical datasets.

\subsubsection*{Existing Approaches}
Within these body modality methods there are two primary approaches: body expression categorization
and hand-crafted distress behaviour descriptors.
The first approach uses unsupervised learning to cluster spatio-temporal descriptors of movement to define
expression codebooks and then uses these codebooks to predict depression severity~\cite{Joshi:uf,Joshi:2013bp,Song:2013ik}.
The second approach defines hand-crafted descriptors of behaviour psychology literature has shown
to correlate to distress~\cite{Fairbanks:1982it}, such as self-adaptors and
fidgeting~\cite{Scherer:2014is,Mahmoud:2013fn}, and to model the intensity of such behaviours
to indicate distress severity.

\subsubsection*{Thesis Contributions}
My core research question is: \textbf{is meta information from gestures predictive of psychological distress?}

In this dissertation I present novel generic \textbf{body gesture meta features} for automatic
distress detection.
These features are based on the movement, frequency, and duration of body gestures.
Per-gesture these features are cheap to compute and relatively simplistic, aggregated across all
gestures they aim to represent general behavioural characteristics of a subject.

I evaluate the use of these features for detecting depression (non-clinical) and find that they
provide useful predictive information within linear models.
I then evaluate the generalisability of these features by applying them to classification tasks for
other distress and personality labels.

As I noted earlier, a core barrier to the development of new features within the body modality is
the lack of available distress datasets that expose the required source data.
As such I introduce a novel audio-visual dataset gathered for this dissertation in collaboration
with Dr. Marwa Mahmoud.
The dataset contains audio-visual recordings of semi-structured interviews and labels based on established
psychology self-evaluation questionnaires for distress and personality measures.

\section*{Dissertation structure}
Chapter~\ref{ch:literature-review} contains a literature review covering related distress detection
research and relevant technology for working with the body modality specifically.
Chapter~\ref{ch:dataset} introduces a novel dataset for depression detection.
I present my method in Chapter~\ref{ch:method}.
I evaluate my features' validity in Chapter~\ref{ch:results}.
Finally, I conclude and outline potential future work in Chapter~\ref{ch:conclusion}.

\chapter{Literature Review}\label{ch:literature-review}
This review is structured in two parts: automatic depression detection and body modality technology.
In the first component I review existing approaches to automatic depression
detection and related fields, and the variety of modalities used by these approaches.
This defines the research space this dissertation occupies.
The second component outlines the current methods for working with the body modality and the technology
involved in these methods, such as pose estimation.
This provides the basis for the core data I use within my method.

\section{Automatic Detection of Distress}\label{sec:lit-review-predicting-distress}

Distress is expressed through all modalities.
Many approaches have been developed to automatically detect distress using behavioural cues, these include both
mono-modal and multi-modal approaches.
For example, face analysis methods are one of the common, and powerful, techniques enabled by the development of tools for
extracting accurate positional information, such as facial landmarks, and tools for automatic interpretation
based on existing psychology approaches such as FACS~\cite{Ekman:Rzx-k0OT}.

I review uses of the primary modalities, cross-modal considerations, multi-modal approaches, and
the expanding use of deep learning within automatic distress detection.
The face, head, and body modalities are the most relevant, though I briefly provide examples of
text and vocal modal usage.

\subsection*{Text Modality}

Dang et al.\ \cite{Dang:2017fd} use linguistic attributes, auxiliary speech behaviour, and word
affect features to predict depression severity and emotional labels within the DAIC dataset~\cite{Gratch:2014vo}.
Linguistic attribute features include total number of words, unique words, pronouns, lexical proficiency,
among others.
Auxiliary speech behaviour cover parallel actions, such as laughing or sighing, and meta information
such as word repeats and average phrase length.
Word features are determined by a collection of annotated corpora that assign $n$-grams categorizations
or ratings relating to their affective semantic.
For example, assigning an emotion type (anger, disgust, joy, etc) to a word or rating words from 0
to 10 on affective attributes such as arousal, valence, dominance, and pleasure.
These three feature types are used to predict depression measures.

\subsection*{Audio Modality}
The audio modality can be a strong predictive source as non-verbal features of speech can be predictive of
distress irrespective of the content of a person's speech~\cite{Ozdas:do}.

Features from audio data commonly include prosody, jitter, intensity, loudness, fundamental frequency,
energy, Harmonic-to-Noise-Ratio (HNR), among others.
Ozdas et al.\ \cite{Ozdas:do} present a method for analysing fluctuations in the fundamental frequency
of a person's voice to assess their risk of suicide.
Alghowinem et al.\ \cite{Alghowinem:2016jq} explore the use of a broad selection of vocal
features, extracted using the \enquote{openSMILE} toolkit~\cite{Eyben:2010fq} to detect depression.
Dibeklioglu et al.\ \cite{Dibeklioglu:2015cf} use vocal prosody to detect depression.

In their investigation of psychomotor retardation caused by depressive states,
Syed et al.\ \cite{Syed:2017jb} use low-level descriptors to model
turbulence in subject speech patterns.
By profiling the turbulence of depressed and non-depressed participants with a depression dataset
they develop a model for predicting depression severity based on the level of turbulence.

\subsection*{Facial Modality}

Joshi et al.\ \cite{Joshi:uf} present a \enquote{bag of facial dynamics} depression detection method
based on the same expression clustering as their \enquote{bag of body dynamics} method described in
more depth below.
For this facial method the space-time interest points (STIPs) are generated for face aligned versions of the
source videos.

While Joshi et al.\ use a categorization approach, Dibeklioglu et al.\ \cite{Dibeklioglu:2015cf} use
generic features representing facial movement dynamics for depression detection.
This method involves generating statistical derivations from movement features such as velocity,
acceleration, and facial displacement over a time period and then modeling their effect.

Whilst Dibeklioglu et al.\ take a generic approach to facial movement, Syed et al.\ \cite{Syed:2017jb}
attempt to model behaviour discussed in psychology literature, using features representing psychomotor
retardation to predict depression severity.
Psychomotor retardation has been shown to be linked to depression~\cite{Sobin:wr}.
In particular, Syed et al.\ generate features to capture craniofacial movements that represent
psychomotor retardation, and thus indicate depression.
To capture the target movements they design features that represent muscular tightening, a depressed
subject is expected to have impaired muscle movements.
They model three types of movement: head movement, mouth movement, and eyelid movement.
These movements are represented by temporal deltas to define the amount of movement in the region.
From these localised movement deltas the authors aim to represent specific actions such as blinking
or contorting of the mouth.
Relatively simplistic features derived from these actions, such as blink rate, can be indicative
of depression~\cite{Alghowinem:gca,Ebert:1996be}.
The nature of human behaviour is that these kinds of simplistic features can contribute useful
information to distress detection models.
Moreover, modeling specific facial actions has been examined as well.
For example, Scherer et al.\ \cite{Scherer:2014is}
use smile features such as intensity and duration, along with other modalities, to detect depression.

Yang et al.\ \cite{LeYang:2017kc} present a novel facial descriptor, a
\enquote{Histogram of Displacement Range (HDR)}, which describes the amount of movement of facial landmarks.
The histogram counts the number of occurrences of a displacement within a certain range of movement.
Where Syed et al.\ represented the amount of movement of certain facial features to measure psychomotor
retardation, Yang et al.\ represent the number of times the face is distorted, so to speak, by landmarks
moving a certain amount.

\paragraph{Eye Sub-Modality}
While Syed et al.\ \cite{Syed:2017jb} explored the use of eye lid features, eye gaze features have
also been shown to be effective in predicting distress.
This modality has become viable as eye tracking technology has progressed sufficiently to enable
accurate processing of eye features.

Alghowinem et al.\ \cite{Alghowinem:2015cs} use eye gaze/activity to perform binary
classification of depression in cross-cultural datasets.
They extract iris and eyelid movements to extract features such as blink rate, duration of
closed eyes, and statistical \enquote{functionals} (i.e.\ simple derivations) of the amount of
activity.
However, activity is not the only indicator, Scherer et al.\ \cite{Scherer:2014is} use average eye gaze
vertical orientation (among other features), in the span $[-60,60]$ degrees.

\subsection*{Head Modality}
Joshi et al.\ \cite{Joshi:uf} present a \enquote{histogram of head movements} depression detection
method that models movement of a person's head over time.
They use three facial landmarks, the corner of each eye and the tip of the nose, to compute the
orientation of the subject's head.
The histogram uses orientation bins of width 10 within the range $[-90,90]$ degrees for windows of
time within a video.
These windowed histograms are then averaged over the full length of the video.
The resulting average histogram is a descriptor of the amount of movement within the video by
representing the variety of angles the head orients to.
This method achieves comparable performance to their \enquote{bag of facial dynamics} method.

A number of the methods using eye activity features also incorporate head activity in their models.
Alghowinem et al.\ \cite{Alghowinem:2015cs} model head activity similarly to their modeling of
eye activity.
As with eye activity, they extract statistical derivations of movement and angular shift.
They also include the duration of the head at different orientations, the rate of change
of orientation, and the total number of orientation changes.
Scherer et al.\ \cite{Scherer:2014is} use head vertical orientation, similar to their eye gaze
orientation feature, as a feature for predicting depression.
They use the average pitch of the head within a 3D head orientation model.
Dibeklioglu et al.\ \cite{Dibeklioglu:2015cf} also model head movement dynamics in a similar fashion
to their facial movement dynamics features.
Similar to Alghowinem et al.\ they extract statistical derivations of movement velocity, amplitude,
and acceleration as head movement features.

The head movement statistical derivation features presented by Alghowinem et al.\ and Dibeklioglu et al.\
are similar to the features I introduce in this dissertation, in that they represent meta movement
information of the modality, rather than categorizing movement.
Though, of course, Alghowinem et al.\ also incorporate categorized movement via their orientation
features.

\subsection*{Body Modality}
The body modality is the least researched of the modalities reviewed.
This is due to a number of factors including the difficulty of dataset creation and the available
technology for extracting raw body data.
Contrast this with the relative ease of working with the other modalities and it is no surprise
they received more attention.
However, much of the relevant body modality based automatic distress detection research appeared in the early 2010s
as some private clinical datasets were gathered and the parallel technological ecosystem expanded
to support generation of body modality features.

\subsubsection*{The Case for Further Investigation}
De Gelder~\cite{deGelder:2009bz} presents the case for further research of bodily expressions
within affective neuroscience.
Though a parallel field, the core argument is much the same for affective computing's investigation of
bodily expressions.
Specifically, at the time of writing (2009) de Gelder asserts that 95\% of \enquote{social and affective
neuroscience} focuses on faces and that the remaining 5\% is mostly split between vocal, musical, and environmental
modalities with a very few number of papers investigating the body modality.
This was similar to the state of automatic distress detection in the early 2010s, though the vocal
modality has a more prominent position in the literature and is more evenly
balanced with the facial modality.

\paragraph{Affectation control and robust automatic detection}
Non-verbal features provide discriminative information regardless of a person's conscious
communication, this is particularly important for automatic distress detection.
Different modalities can be consciously controlled to varying degrees.
For example, facial expressions are more easily controlled than bodily expressions~\cite{deGelder:2009bz}.
By including more modalities, and representations of those modalities, automatic distress detection
could become more robust to conscious affectation modification of modalities.

Further to robustness, one of the advantages of the face and body modalities is their ability to detect
micro-expressions.
Micro-expressions are instinctual reactions to some stimulus that can be predictive of emotional
and distress state~\cite{Ekman:N3183wDU,Haggard:1966ct}.
They are significantly harder to control than general expressions.

\subsubsection*{Expression Categorization}
Much of the body modality research has approached the problem as a transfer of the methods from the facial
modality by modeling body movements as expressions in much the same way facial expressions are~\cite{Joshi:2012uma,Joshi:uf,Joshi:2013bp}.
Differences in these methods have been centred around: what tracklets form the basis of the body data~\cite{Joshi:uf},
the use of deep learning vs manual descriptor definition~\cite{Scherer:2014is}, and process for generating categories
of expressions~\cite{Song:2013ik}.

Joshi et al.\ \cite{Joshi:uf} demonstrate the discriminative power of bodily expressions for predicting
clinically based depression measures.
They use STIPs from recordings of a participant's \textit{upper} body
and generate a \enquote{Bag of Body Dynamics (BoB)} based on codebooks of expression representations.
STIPs are generated for a video, then Histograms of Gradient (HoG) and Optical Flow (HoF) are computed
spatio-temporally around the STIPs, these histograms are then clustered within a sample, the cluster centres
form a representation of the movements occurring in the video.
The cluster centres from all videos in a training set are clustered again to generate the codebook
of expressions.
A BoB feature vector is generated for each sample by counting the number of
cluster centres within the sample that fit within each codebook expression.
Finally, these BoB feature vectors are used by an SVM to detect depression.

Joshi et al.\ \cite{Joshi:2013bp} extend on this approach by combining a \enquote{holistic body
analysis}, similar to the BoB method, this method uses STIPs for whole body motion analysis and adds
relative body part features.
These features represent the movement of the head and limbs relative to the participant's trunk,
represented as polar histograms.

Applying the same essential method as Joshi et al.,\ Song et al.\ \cite{Song:2013ik} present a method
for learning a codebook of facial and bodily micro-expressions.
They identify micro-expressions by extracting STIPs over very short time intervals (e.g.\
a few hundred milliseconds), then, as Joshi et al.\ do, they compute local spatio-temporal features
around the STIPs and learn a codebook of expressions based on these local features.
Finally, for each sample they generate a Bag-of-Words style feature vector based on the
codified micro-expressions present in the sample.

\subsubsection*{Distress Behaviour Descriptors}

Psychology literature describes specific behaviours that are correlated with psychological distress
and disorders.
For example, Fairbanks et al.\ \cite{Fairbanks:1982it} find self-adaptors and fidgeting behaviour
to be correlated to psychological disorders.
Based on this work, Scherer et al.\ \cite{Scherer:2014is} evaluate the use of these behaviours
for automatic distress detection.
Specifically, they manually annotate their dataset for hand self-adaptor behaviours and fidgets,
including hand tapping, stroking, grooming, playing with hands or hair, and similar behaviours.
To identify whether regions are relevant to these behaviours they also annotate these behaviours
with categories such as head, hands, arms, and torso, and then extract statistical information
such as the average duration of self-adaptors in each region.
They also annotate leg fidgeting behaviour such as leg shaking and foot tapping, again they use
statistical derivations of these behaviours as features for detection.

Whilst Scherer et al.\ manually annotated self-adaptors and fidgets, Mahmoud et al.\ \cite{Mahmoud:2013fn}
present an automatic detector of fidgeting, and similar behaviours, based on a novel rhythmic motion
descriptor.
They extract SURF interest point tracklets from colour and depth data and then apply their novel rhythmic
measure to check similarity among cyclic motion across tracklets.
Rhythmic motion is then localised based on Kinect skeletal regions and classified as one of four
classes: Non-Rhythmic, Hands, Legs, and Rocking.

\subsection*{Multi-Modal Fusion}
Combining modalities for prediction has proven effective when combining a variety of
modalities~\cite{LeYang:2017kc,Joshi:2013cp,Dibeklioglu:2015cf}.
There are four primary types of fusion: feature fusion such as feature vector concatenation,
decision fusion such as majority vote, hybrid fusion which uses both, and deep learning fusion
which merges inner representations of features within a deep learning architecture.
The deep learning fusion method differs from feature fusion as the features are provided
to separate input layers and only merged after inner layers, but before decision layers.

Song et al.\ \cite{Song:2013ik} combine micro-expressions from facial and bodily modalities with
sample-level audio features.
They evaluate three methods, early fusion by concatenating audio features to features from each
visual frame, early fusion using a CCA~\cite{Hardoon:2006gp} kernel, and late fusion based on voting, where the
per-frame predictions from the visual modalities are averaged over the sample and then combined with
the audio prediction.
Dibeklioglu et al.\ \cite{Dibeklioglu:2015cf} fuse facial, head, and vocal modalities using
feature concatenation.
They extend on this by performing feature selection on the concatenated vector, rather than the source
vectors, using the Min-Redundancy Max-Relevance algorithm~\cite{Peng:ta}.

Alghowinem et al.\ \cite{Alghowinem:2015cs} perform hybrid modality fusion, both combining feature
vectors from modalities and performing a majority vote on individual modality classification predictions.
The vote fusion is based on three classifiers, two mono-modal classifiers and one feature fusion classifier.

Huang et al.\ \cite{Huang:2017er} train long-short-term memory (LSTM) models on facial, vocal, and text modalities
and then use a decision level fusion, via a SVR, to predict the final regression values.
This paper differs from many deep learning approaches as it uses the decision
level fusion, rather than having the deep learning models find patterns across feature types.

\paragraph{Temporal contextualisation}
Gong \& Poellabauer~\cite{Gong:2017fj} present another approach to usage of multiple modalities where
the text modality provides contextualisation for features from the audio-visual modalities.
They apply a topic modeling method for depression detection using vocal and facial modalities,
where features are grouped based on the topic being responded to within an interview.
The authors suggest that without topic modeling the features are averaged over too large a time period
such that all temporal information is lost.
By segmenting the samples they aim to retain some of the temporal information.
Arbitrary segmentation would not necessarily be useful, thus their use of logical segmentation based
on topic.

\subsection*{Deep Learning}

Much of the recent work in distress detection leverages advances in deep learning, especially
advances related to recurrent architectures, such as LSTMs,
which can model sequence data well.
In distress detection most modalities provide sequence data, audio-visual streams, natural language,
or sequence descriptors of the data (such as FACS AUs).

Chen et al.\ \cite{Chen:2017ca} present a method utilizing text, vocal, and facial modalities for
emotion recognition.
They explore the use of existing vocal features such as fundamental frequency analysis, auto-learnt features based on
pre-trained CNNs to extract vocal and facial features, and word embedding features for the text.
The auto-learnt facial features are derived from existing facial appearance data already extracted from
the raw videos (i.e.\ they do not have their CNNs process raw frames to extract features).
They then experiment with SVR and LSTM models to evaluate the temporal value of the LSTM model.
They find that fused auto-learnt features from all modalities in combination with the LSTM model
provides the greatest performance.

Yang et al.\ \cite{LeYang:2017ir} present an interesting multi-level fusion method that incorporates
text, vocal, and facial modalities.
They design a Deep Convolutional Neural Network (DCNN) to Deep Neural Network (DNN) regression model that is
trained, separately, for audio and video modalities.
They also trained their regression models separately for depressed and non-depressed participants,
resulting in four separate DCNN - DNN models.
They use the openSMILE toolkit for their audio features, this is common among many of the vocal modality
methods (e.g.\ Alghowinem et al.\ from above), and FACS AUs for their visual features.
They derive a temporally relevant feature vector from the set of all AUs by calculating the change
in AUs over time.
Their text modality model uses Paragraph Vectors in combination with SVMs and random forests
and predicts a classification task rather than a regression task.
Finally, they fuse, using DNNs, the audio and visual model predictions per training set, i.e.\ the depressed
participant trained models are fused and the non-depressed models are fused.
They then use another DNN to fuse the two fused regression predictions (i.e.\ depressed and non-depressed)
and the classification prediction from the text modality.
While they use DNNs to fuse decisions at multiple levels, this is a decision-level fusion method,
not a deep learning fusion method, as they fuse the regression predictions from each model rather than
the inner layer outputs.

Yang et al.\ \cite{LeYang:2017kc} present a second paper which utilises the same structure of DCNNs
and DNNs, with two significant changes: firstly, the video features are changed to a new global
descriptor they present, the \enquote{Histogram of Displacement Range (HDR)} which describes the
amount of movement of facial landmarks, and secondly, the text
modality now uses the same DCNN - DNN architecture to perform regression based on text data.
Having changed the output of the text model the final fusion is a regression fusion using the same
method as the audio visual models were fused in the first paper.

Finally, no method, that I am aware of, uses deep learning end-to-end for automatic distress detection
such that features are learnt from raw data for predicting distress.
All methods apply deep learning on top of audio-visual descriptors and hand-crafted features.
One of the core difficulties is the relative sparsity of existing data and thus the restricted
ability to learn interesting features.
Therefore, continued development of features and approaches to behavioural representation is valuable
and able to contribute to methods that use deep learning.

\section{Body Gestures}\label{sec:lit-review-body-gestures}
Body gestures present a number of challenges including deriving features directly from pixels, extracting
positional data, detecting gestures within the data, and designing representative features based on
gestures.
This section focuses on previous work on body gesture and pose representation, which forms the basis
of my feature definition and generation.

There are three primary approaches to body gesture representation within the affective computing literature:
traditional computer vision feature detection algorithms such as STIPs~\cite{Joshi:uf,Joshi:2013bp},
pose estimation~\cite{Cao:2018tk} from standard video recordings,
and use of specialised 3D capture equipment such as Kinects~\cite{Mahmoud:2013fn}.

\subsection*{Recording Type - 2D vs 3D}
The first two approaches use standard 2D video recordings to extract the base data for calculating features.
The third approach uses 3D video recordings to enable more detailed analysis of body movement via
depth.
The most common 3D capture equipment used within related work is Kinects.
Kinects have an added benefit that they provide skeletal key points (i.e.\ joint locations) within
a recording, thus enabling more accurate representation of body data.

\subsection*{Extracting Body Representations}
Given a video recording, 2D or 3D, with or without skeletal information, the next phase is feature
extraction.
Feature extraction approaches fall into two primary categories, which apply to both 2D and 3D videos:
generic video feature representations and body modality specific representations.

\subsubsection*{Generic Interest Points}
The first approach does not target specific body areas or movement, instead it assumes that the only
subject within the video is the subject the model is concerned with, i.e.\ the participant, and
extracts generic video features from the recording to represent the body and gestures.
Examples of these generic features are Space-Time Interest Points (STIPs), SURF, Histogram of Gradients
(HOGs) and Optical Flow (HOFs), among others.
These features can be extracted from colour and depth recordings such that they are applicable to both
2D and 3D recordings.

While some approaches apply these generic features (or rather, derivations of these features) directly to
prediction tasks~\cite{Joshi:uf,Song:2013ik} (examples are discussed in Section~\ref{sec:lit-review-predicting-distress}),
others aim to incorporate heuristics and information based on body parts.
For example, interest points can be segmented based on location within the frame to heuristically distinguish
body parts (e.g.\ head and arms)~\cite{Joshi:2013bp}.

Another example is Mahmoud et al.\ \cite{Mahmoud:2013fn} who use SURF keypoints from colour and depth
images across a video to define tracklets for their rhythmic motion measure.
Their dataset is based on Kinect captured videos to provide depth, this also provides skeletal regions
such as feet and head.
They use these Kinect skeletal regions to localise their keypoints' motion.

\subsubsection*{Body Modality Specific}
The second approach extracts body modality specific interest points (e.g.\ skeletal models) to calculate
features from.
There are two primary methods for extracting these interest points: joint estimations from specialised
capture equipment such as Kinects and pose estimation from existing video recordings.
Such skeletal models have gained popularity in the past few years for action recognition
tasks~\cite{Su:2017du,Du:2015ei,Chen:2017fs,Shukla:2017gn,Wei:2017he,Du:2015ht}.

In this dissertation I use pose estimation to extract body interest points (e.g.\ joints) from
each frame in a two-dimensional video.
In the past three to four years there has been substantial work on these pose estimation
systems~\cite{Wei:2016io,Cao:2016vv,Simon:2017vl,Zhang:2016gz,Liu:2015ks}.
The current state-of-the-art is OpenPose by Cao et al.\ \cite{Cao:2018tk}.
OpenPose uses skeletal hierarchy and part affinity fields to estimate pose interest points (i.e.\ joints)
relative to each other and determine both part and orientation based on the direction of the proposed
limb.
The state-of-the-art model, at the time of writing, generates 25 interest points identifying the location
of all major joints and more detailed information regarding head and feet.
OpenPose can also be used to extract facial landmarks and detailed hand models.

\section{Summary}\label{sec:literature-review-summary}
I have reviewed methods for automatic depression detection and the variety of modalities and models
used within the field.
I have also discussed, in broad terms, methods used for working with body data within related fields,
from variation in capture equipment to difference in core interest points.

The first component of the review outlined the gap within the existing use of the body modality that
my proposed methodology addresses.
While the second component outlined methods for working with the body modality, and specifically
technology that I use to implement my methodology.

\chapter{Dataset}\label{ch:dataset}

To develop whole body features I need a dataset containing video recordings and distress labels.
However, due to the sensitive nature of this data no existing dataset makes both the source video
and distress labels publicly available.
Though some datasets make video derived data available, such as facial landmarks, none include derived
data for the participant's whole body.
Given this I introduce a new dataset containing audio-visual recordings of interviews and labeled
with distress and personality measures.

This is not a clinical dataset, the interviews are performed by a computer science researcher and
the labels are based on established psychology self-evaluation questionnaires.

\subsection*{Collaborators}
Dr. Marwa Mahmoud from the Department of Computer Science at the University of
Cambridge was the interviewer.
Dr. Gabriela Pavarini from the Department of Psychiatry at the Univeristy of Oxford provided advice
regarding the interview procedure for collection.

\section{Existing Related Datasets}\label{sec:existing-datasets}

There are a number of relevant existing datasets, however none satisfy the requirements of my research.
Namely that they: are available, include psychological distress labels, and include source videos.

I provide an overview of four existing datasets, two clinical datasets, a distress focused dataset
using facial, auditory, and speech modalities, and one affective computing body gesture dataset.
Each of these datasets, and their related work, provide useful insight for my data collection and
model development research.

\subsection{Clinical}\label{subsec:clinical-databases-at-cmu}

Joshi et al.\ describe a clinical dataset~\cite{Joshi:2012uma} for depression analysis that contains
recordings of participants' upper bodies and faces during a structured interview.
It includes clinical assessment labels for each participant, including depression severity.
This dataset was collected at the Black Dog Institute~\cite{blackDogInst}, a clinical research institute in Australia.

Joshi et al.\ \cite{Joshi:2013bp} describe a clinical dataset collected at the University of Pittsburgh
containing full body recordings of participants in interviews, it uses Major Depressive Disorder (MDD)~\cite{Association:1994te}
diagnosis and Hamilton Rating Scale of Depression (HRSD)~\cite{psychology:ui} as labels.
The authors validate the use of full body features for depression prediction within this dataset.

Given the clinical nature of these datasets they are not publicly available.

\subsection{Audio-Visual Distress}\label{subsec:daic-woz}

Gratch et al.\ introduce the Distress Analysis Interview Corpus (DAIC) dataset~\cite{Gratch:2014vo}
containing audio-visual recordings of interviews with participants with varying levels of distress.
Participants are assessed for depression, post traumatic stress disorder (PTSD), and anxiety based
on self-evaluation questionnaires.
The dataset provides audio data (and derivations such as transcripts) and extracted facial landmark
data, though they do not provide source video data nor pose data.
Source video recordings are rarely shared in distress focused datasets due to the sensitive of the
data.

The dataset contains four participant interview structures: face-to-face interviews with a human,
teleconference interviews with a human interviewer, ``Wizard-of-Oz'' interviews with a virtual agent
controlled by an unseen interviewer, and automated interviews with a fully autonomous virtual agent.

This dataset informs my dataset collection via the labels it contains and their interview structures.
The DAIC method has participants complete a set of questionnaires, then participants are interviewed
and recorded, and finally the participant completes another set of questionnaires.
Within the interview the interviewer asks a few neutral questions to build rapport, then asks questions
related to the participant's distress symptoms, and finally asks some neutral questions so the
participant can relax before the interview finishes.

\subsection{Body Gestures}\label{subsec:body-gesture-datasets}
Palazzi et al.\ introduce a audio-visual dataset of dynamic conversations between different ethnicities
annotated with prejudice scores.
Videos in this dataset contain two participants interacting in an empty confined space.
Participants move around the space throughout the session, enabling analysis of body language
affected during conversation.

This dataset highlights, and focuses on, the effect attributes of a counterpart, such as race,
gender, and age, have on a person's behaviour.
In this dataset counterparts are explicitly non-uniform.
Whereas, in my dataset the interviewer (i.e.\ counterpart for a participant) is the same for all interviewees.
This is useful in controlling for the counterpart variable in human behaviour, supporting the
isolation of correlations between distress and behaviour.

Datasets such as this one are not useful for my current research question, however, as they provide
source videos they present opportunities for future work investigating my features generalisability
to other domains.

\section{Method}\label{sec:dataset-method}

\subsection{Design}\label{sec:design}
This dataset is designed to enable investigation of the body modality for use in automatic detection
of distress for early screening.
This is a non-clinical dataset.

Its source data is audio-visual recordings of conversational interviews.
The recordings capture the whole body of participants to enable features based on a whole body modality.
These interviews involve questions related to distress to elicit emotive responses from participants,
however the responses to these questions are irrelevant to the core data.
The interviews use a conversational style to best enable naturalistic gestures from participants.

Labels are scored results from established self-evaluation questionnaires for assessing distress
and personality traits, as well as demographic labels such as gender.
The distress questionnaires are: the PHQ-8~\cite{Kroenke:2009ky, Kroenke:2001fl} for depression,
GAD-7~\cite{Spitzer:2006gb} for anxiety, SSS-8~\cite{Gierk:2014fv} for somatic symptoms,
and the PSS~\cite{PerceivedStressSca:1983dv} for perceived stress.
Personality traits are measured using the Big Five Inventory~\cite{John:ud}.

\subsection{Participants}\label{subsec:dataset-participants}

\subsubsection*{Recruitment}

I advertised for participants via University of Cambridge email lists, student social media groups,
classified sections of websites, such as Gumtree~\cite{gumtree}, specific to the Cambridge area, and paper
fliers posted around the University of Cambridge.
Participants were directed to a website\footnote{\url{helpresearch.org}} describing the study
along with a participant registration form.

The registration form captured demographic data and two self-evaluation psychological distress
questionnaires.
Demographic data captured includes gender, age, ethnicity, and nationality.
Gender and age were required while ethnicity and nationality were not.
The two psychological distress questionnaires were the PHQ-8~\cite{Kroenke:2009ky, Kroenke:2001fl}
and GAD-7~\cite{Spitzer:2006gb}.

\subsubsection*{Selection}

In total 106 people registered to participate and 35 were invited to the face to face session.
The participant population is balanced with regards to distress levels and gender\footnote{
Non-binary/other was given as an option in the registration form.
A number of people registered with this option.
However, none of those people met the distress level criteria and were thus not selected for an interview.}.
Distress level balancing aims to include participants at the extents of the distress spectrum such
that there is a distinct difference between the high and low distress populations.
Participant distress level is selected based on PHQ-8 and GAD-7 questionnaire responses such that
participants are balanced between high (i.e.\ major or severe) and low (i.e.\ mild) distress.
Of the invited participants, there are 18 with high distress and 17 with low distress.

\subsubsection*{Compensation}
Participants are compensated for their time with a \pounds15 voucher.

\subsection{Face to Face Session}\label{subsec:process}

During the face to face session participants sign a research consent form outlining the interview
process, complete a battery of five established psychology questionnaires evaluating
distress levels and personality traits, are interviewed by a researcher, and finally sign a debrief
research consent form that outlines the full purpose of the study.
Participants are not aware of the focus of the research (i.e.\ body modality analysis) before the
interview such their affectations are natural.

To achieve the conversational interview dynamic the interviewer asks general questions regarding the
participant's life and further encourages the participant to elaborate.
For example, the interviewer asks \enquote{can you tell me about one time in your life you were particularly
happy?\@} and then asks follow up questions regarding the example the participant provides.
The interview style and structure is inspired by the DAIC dataset.
In developing the interview structure and questions I also drew on training documents provided by
Peer2Peer Cambridge~\cite{peer2peer}, an organisation dedicated to peer support for mental health
which trains students in general counseling.

So as to avoid influencing participants' behaviour the interviewer remains as neutral as possible
during the interview, while still responding naturally such that the participant is comfortable
in engaging in the questions.
Furthermore, to ensure neutrality the interviewer is explicitly not aware of the distress level of participants
before the interview and has no prior relationships with any participant.

\subsubsection*{Technical Faults}
16 interviews are interrupted due to technical faults\footnote{The camera disconnected.}.
Recordings that are interrupted are treated as multiple individual samples within the dataset
(though they remain connected by their participant ID).

\section{Preliminary Analysis}\label{sec:dataset-analysis}

The dataset contains a total of 35 interviewed participants with a total video duration of 7 hours
50 minutes and 8 seconds.
Each participant provides responses to 5 questionnaires, including 2 responses to both the PHQ-8 and
GAD-7 questionnaires as participants completed both during registration and the face-to-face session.

Though significantly more people registered for participation, I include only those interviewed
in this analysis.

\subsection{Validation Criteria}\label{subsec:dataset-statistics}

There are three primary criteria the dataset should satisfy with regards to label results:
\begin{enumerate}
  \item The psychological measures statistically match previous work and published norms\footnote{
  Published norms are the standard values for questionnaire results as defined by the psychology literature.
  These aim to be representative of the general population.
  They thus provide a benchmark for other work (generally within psychology) to check smaller populations'
  results against.} (i.e.\ the
  distribution within the participant population is similar to that of the general population).
  \item There are no confounding correlations.
  For example, gender correlating highly to depression
  would indicate a poorly balanced dataset and would be confounding for depression analysis.
  \item The labels are well balanced to enable machine learning.
\end{enumerate}

As the common measure of similar distress detection research is depression I focus primarily on it
for this validation.

General statistics regarding the questionnaire and demographic results within the dataset are provided
in Table~\ref{table:dataset-primary-statistics}.
Covariance is presented as normalized covariance values, also known as the correlation coefficient.
\begin{table}[h]
  \small
  \begin{center}
    \begin{tabular}{rrrrrrrr}
      \toprule
      \textbf{Label} &
      \makecell{\textbf{Possible} \\ \textbf{range}} &
      \textbf{Max} &
      \textbf{Min} &
      \textbf{Mean} &
      \textbf{Median} &
      \textbf{Std.} &
      \makecell{\textbf{Depression} \\ \textbf{covariance}}
      \\ [0.5ex]

      \midrule
      \makecell[l]{\textbf{Distress}} &&&&&&&  \\
      Depression & 0--24 & 19 & 0 & 7.43 & 8 & 5.87 & - \\
      Anxiety & 0--21 & 19 & 0 & 7.00 & 8 & 5.53 & 86.15\% \\
      Perceived stress & 0--40 & 30 & 1 & 18.17 & 18 & 8.03 & 84.00\% \\
      Somatic symptoms & 0--32 & 27 & 1 & 9.06 & 7 & 6.94 & 74.16\% \\
      \midrule
      \makecell[l]{\textbf{Personality}} &&&&&&&  \\
      Extraversion & 0--32 & 31 & 3 & 16.37 & 17 & 6.33 & -30.49\% \\
      Agreeableness & 0--36 & 34 & 12 & 25.67 & 26 & 5.60 & -42.21\% \\
      Openness & 0--40 & 39 & 7 & 27.29 & 28 & 6.77 & 4.29\% \\
      Neuroticism & 0--32 & 31 & 1 & 16.86 & 18 & 8.60 & 80.00\% \\
      Conscientiousness & 0--36 & 36 & 10 & 21.46 & 21 & 6.87 & -46.41\% \\
      \midrule
      \makecell[l]{\textbf{Demographic}} &&&&&&&  \\
      Gender & - & - & - & - & - & - & 9.47\% \\
      Age & - & 52 & 18 & 25.40 & 22 & 9.1 & -11.09\% \\

      \bottomrule
    \end{tabular}
  \end{center}
  \caption{
  General statistics regarding questionnaire and demographic results within the dataset.
  The \enquote{Depression covariance} column is most important as it demonstrates the independence
  of variables with regards to depression (for example, it shows that age and gender are not
  confounding of depression).
  }
  \label{table:dataset-primary-statistics}
\end{table}

\subsubsection*{Published Norms}
A comparison of the mean values for distress and personality measures between my dataset and the
published norms is presented in Table~\ref{table:dataset-published-norms}.
While there are differences, the measures are generally in line with the published norms.
The dataset has slightly higher mean distress scores, though a substantially higher mean perceived stress
score.
Depression, extraversion, and neuroticism measures are particularly close to their published norms.
While the dataset mean for agreeableness and openness are substantially greater than the published
norms (over 10\% over the technical range for those measures).

\begin{table}[h]
  \small
  \begin{center}
    \begin{tabular}{rrrr}
      \toprule
      \textbf{Label} &
      \textbf{Dataset mean} &
      \textbf{Norm mean} &
      \textbf{Source}
      \\ [0.5ex]

      \midrule
      \makecell[l]{\textbf{Distress}} &&&  \\
      Depression & 7.43 & 6.63 & Ory et al.\ \cite{Ory:2013bs} \\
      Anxiety & 7.00 & 5.57 & Spitzer et al.\ \cite{Spitzer:2006gb} \\
      Perceived stress & 18.17 & 12.76 & Cohen et al.\ \cite{PerceivedStressSca:1983dv} \\
      Somatic symptoms & 9.06 & 12.92 & Gierk et al.\ \cite{Gierk:2014fv} \\
      \midrule
      \makecell[l]{\textbf{Personality}} &&& \\
      Extraversion & 16.37 & 16.36 & Srivastava et al.\ \cite{Srivastava:tc} \\
      Agreeableness & 25.67 & 18.64 & Srivastava et al.\ \cite{Srivastava:tc} \\
      Openness & 27.29 & 19.61 & Srivastava et al.\ \cite{Srivastava:tc} \\
      Neuroticism & 16.86 & 16.08 & Srivastava et al.\ \cite{Srivastava:tc} \\
      Conscientiousness & 21.46 & 18.14 & Srivastava et al.\ \cite{Srivastava:tc} \\

      \bottomrule
    \end{tabular}
  \end{center}
  \caption{
  Comparison of the mean questionnaire values within my dataset to the published norms.
  This shows that the population distribution, with regards to these distress and personality measures,
  is generally in line with the broader population.
  }
  \label{table:dataset-published-norms}
\end{table}

\subsubsection*{Confounding Correlations}
While the other distress measures (anxiety, perceived stress, and somatic stress) are strongly
correlated with depression,
the personality measures have below 50\% covariance with the exception of neuroticism which has
an 80\% covariance.
Furthermore, the demographic measures, gender and age, are negligibly correlated, with 9.47\% and
-11.09\% covariance, respectively.
This suggests that the labels are not confounding of each other.

\subsubsection*{Label Balance}
There are 17 participants below the mean depression result (7.43) and 18 participants above.
The mean depression score of the group below the overall mean is 2.18 while the score for those
above is 12.39.
Ideally for machine learning the dataset's distribution would include more participants at the severe depression
end of the spectrum, though the present distribution still places the below group firmly in the \enquote{mild} category
and the above group in the \enquote{major depression} category.

There are 18 male and 17 female participants.
As the gender covariance shows, the split on the depression measure and the split on gender are not
the same participants (gender is balanced across the distress spectrum).

\subsection{Difference from Registration}\label{subsec:difference-from-registration}
Participants complete the PHQ-8 and GAD-7 questionnaires during registration and during the interview
process.
These questionnaires are temporal, specifically, they relate to the participant's mental state in
the past two weeks.
Given this, some difference between registration and interview results is expected.

With the exception of a small number of outliers, participants were generally consistent
in self-evaluation between registration and interview.
PHQ-8 responses have a mean difference of 0.89 while GAD-7 responses have a mean difference of 0.63.
This supports the selection of participants based on temporal self-evaluation questionnaire results.

\subsection{Interview Meta Statistics}\label{subsec:recordings}

There is a total of 7 hours 50 minutes and 8 seconds of participant interview recordings, with a
mean interview duration of 13 minutes and 25 seconds.
The standard deviation of interview duration is 3 minutes and 20 seconds and the median interview
duration is 13 minutes and 8 seconds.
Depression score and interview duration are not correlated, with a covariance of 6.95\%.
Furthermore, interview duration is not correlated with any questionnaire result (i.e.\
distress or personality measure), all absolute covariance values are below 25\%, which provides
confidence in the reliability of the data.

\section{Summary}\label{sec:dataset-summary}

I have introduced a new audio-visual dataset containing recordings of conversational interviews
between participants and a researcher, and annotated with established psychology self-evaluation
questionnaires for depression, anxiety, somatic symptoms, perceived stress, and personality traits.
This dataset involves 35 participants and 65 recordings (due to recording interruptions) with a
total video duration of 7 hours 50 minutes and 8 seconds.

There are a number of relevant existing datasets including clinical datasets which contain body gestures
but are inaccessible beyond their home institute, distress datasets that contain facial expressions
and speech modalities but no body gestures or source videos, and video datasets containing body
gestures but lacking distress labels.
While these datasets inform my dataset design and collection, no dataset I am aware of satisfies
the criteria for research on body gesture modalities for predicting distress.

An analysis of the questionnaire results in the dataset show they are aligned with psychology literature
published norms, they are not confounded by factors such as gender or age, and have a useful
and balanced distribution across the distress spectrum.

\chapter{Method}\label{ch:method}
\begin{wrapfigure}{r}{0.5\textwidth}
  \vspace{-200pt}
  \begin{center}
    \includegraphics[width=0.48\textwidth]{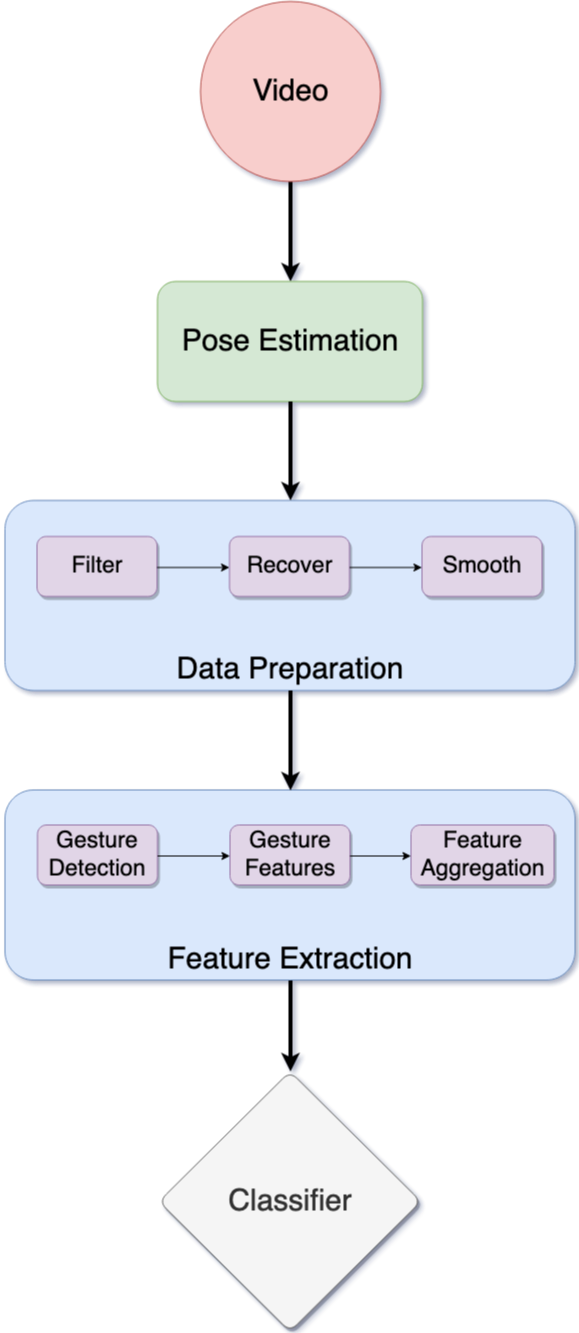}
  \end{center}
  \caption{
  High level pipeline description of feature extraction process.
  }
  \label{fig:method_pipeline}
\end{wrapfigure}

Having collected the dataset, I now describe my methodology for extracting gesture meta features and
applying them to automatic distress detection.
To this end I define four core stages of my methodology, from video to prediction (as shown in
Figure~\ref{fig:method_pipeline}):
\begin{enumerate}
  \item \textbf{Pose estimation} - extract pose estimation data from video recordings.
  \item \textbf{Data preparation} - When preparing the data for learning I clean and smooth the
  extracted pose data.
  Pose data extracted from 2D video frames has a not-insignificant level of noise.
  \item \textbf{Feature extraction} - Detect gestures, extract per-gesture features, and aggregate
  features across all gestures.
  I define gestures as sustained movement of some body part over a minimum period of time (this is
  elaborated on in Section~\ref{subsec:method-features-gestures}).
  \item \textbf{Classifier training} - a classifier is then trained to predict a target label (e.g.\ depression) based on
  these aggregate features.
  Classifier choice is explained in the evaluation, Chapter~\ref{ch:results}.
\end{enumerate}

\section{Pose Estimation}\label{sec:method-pose-estimation}
\begin{figure}[h]
  \begin{center}
    \includegraphics[angle=90,origin=c,height=0.6\linewidth]{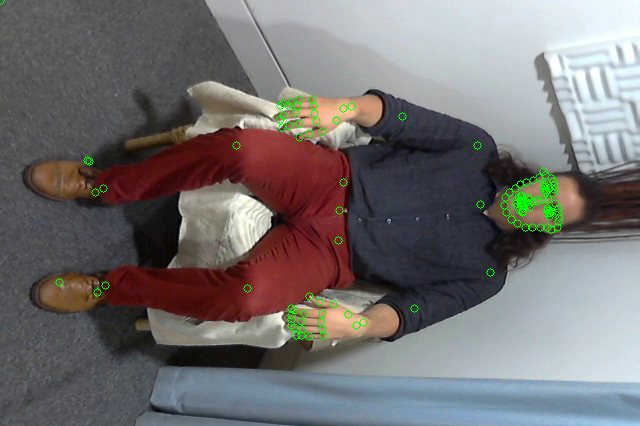}
  \end{center}
  \caption{
  Example of OpenPose estimation output of the participant interview position.
  Subject is not a study participant.
  Pose points are indicated by the green dots.
  }
  \label{fig:pose_point_example}
\end{figure}

Pose estimation extracts per-frame approximations of skeletal joint locations.
This enables more accurate gesture analysis than direct pixel based approaches (such as STIPs
per Joshi et al.'s method~\cite{Joshi:uf}).

I process each video to extract per-frame skeletal pose data using OpenPose~\cite{Cao:2018tk} by
Cao et al.\ (OpenPose is discussed in more detail in Section~\ref{sec:lit-review-body-gestures}) as
it is the current state-of-the-art in pose estimation.
I use the \texttt{BODY\_25} pose model provided with OpenPose\footnote{
OpenPose models provided at \url{https://github.com/CMU-Perceptual-Computing-Lab/openpose}.}.
As the name suggests, this model generates 25 pose points corresponding to a subject's joints.
OpenPose also extracts more detailed hand data that provides joint estimations for each joint in the
hand.
Figure~\ref{fig:pose_point_example} presents an example of extracted pose points.

\section{Data Preparation}\label{sec:data-preparation}
I perform three data preparation steps: filtering, recovery, and smoothing.
Filtering smooths dataset-level noise, recovery fixes outlier noise (where outliers are detection failures
for specific pose points, but not the whole body), and smoothing reduces detection noise.

Extracted pose estimation data has two primary forms of noise: individual frames, or short segments of frames,
where detection of a person, or part of a person, is lost, and detection points moving around slightly
even if the person is static.
Manual review of a selection of detection loss cases shows no consistent cause (for both complete
detection loss and partial loss).
It appears to be the deep learning model failing inexplicably on some frames.

\subsection{Filtering}\label{subsec:filtering}
The absence of, or low representation of, gestures is relevant information for prediction tasks
(simply put, if more activity is relevant then less activity must be too).
However, the absence of gestures can also be due to a lack of opportunity to express gestures,
such as when a sample is too short.
These short samples lead to dataset-level sample noise that can hinder predictive models.

Samples shorter than 1 minute are excluded as it is difficult to provide enough opportunity for gesture
dynamics in less than 1 minute of video.
12 out of 65 samples within the dataset are shorter than 1 minute.

\subsection{Detection Recovery}\label{subsec:detection-recovery}
Pose estimation within my dataset contains frames where certain pose points (e.g.\ an arm or a
leg) are not detected.
In these cases OpenPose returns $0$ for each pose point not detected.
Manual review of a number of instances shows that the joint does not moved much, if at all, during
the lost frames.
However, the pose point \enquote{moving} to position $0$ causes significant noise in feature calculation.

Therefore, I perform detection \enquote{recovery} to infer the position of the pose point in the missing
frames, thus providing a smoother pose point movement.
I recover the position by linearly interpolating the pose point's position between the two closest detected
frames temporally surrounding the lost frame(s).

It is worth noting that this pose point detection failure is different to full detection failure where the
whole participant is not detected within a frame.
I do not attempt to recover such full failure cases as the failure is more serious and cause is
ambiguous.
I do not want to introduce stray data by \enquote{recovering} significantly incorrect data.
Partial failures suggest a simple failure of OpenPose to extend through its body hierarchy.
Since other pose points are still detected I am more confident that it is not a \enquote{legitimate}
failure to do with participant position.
Instead, full failure cases are treated as \enquote{separators} within a sample.
Gesture detection occurs on either side but not across such separators.

\subsection{Detection Smoothing}\label{subsec:detection-smoothing}

To extract more accurate and relevant features I smooth the pose data by removing high frequency
movement within pose points.
Such high frequency movement of pose points is caused by OpenPose's detection noise
(i.e.\ the exact pose point might move back and forth by a few pixels each frame while its target
joint is static).
Thus smoothing is not smoothing human movement, but rather smoothing pose extraction noise.
To smooth the data I apply a fourier transform filter to each dimension for each pose point.
The smoothing steps are:
\begin{enumerate}
  \item Separate the data into $x$ and $y$ positions and smooth them (i.e.\ apply the following steps)
  independently.
  \item I convert the position sequence data using a fourier transform with a window length of $64$.
  \item Set the medium and high frequency values (all frequencies above the first five) to $0$.\
  \item Invert the fourier transform on the updated fourier values to reconstruct the smoothed pose data.
  \item Concatenate the smoothed windows.
\end{enumerate}

\section{Feature Extraction}\label{sec:feature-extraction}

I extract two types of features: aggregate features across the whole body and localised features
for specific body parts, which I term \enquote{body localisations}.
The localisations are: head, hands, legs, and feet.
As the participants are seated the body trunk does not move substantially such that a gesture might
be detected.
The whole-body features include aggregations of the body localisation features as well
as features incorporating the whole body.

By including localised and non-localised features I can model the information provided
by individual body parts and also the overall behaviour.

\subsubsection*{Gesture definition}
I define a gesture as a period of sustained movement within a body localisation.
Multiple body localisations moving at the same time are treated as multiple individual, overlapping,
gestures.

A gesture is represented as a range of frames in which the target body localisation has
sustained movement.

For example, if a participant waves their hand while talking it would be a hand gesture.
If they were to cross their legs it would register as a gesture in both the leg and feet localisations.

\subsubsection*{Whole Body Features}
\begin{itemize}
  \item \textbf{Average frame movement} - the per-frame average movement of every tracked pose point
  (i.e.\ whole body).
  This is the only feature that is not based on detected gestures.
  \item \textbf{Proportion of total movement occurring during a gesture} - the proportion of total
  movement (i.e.\ the whole body) that occurred while some body localisation was affecting a gesture.
  \item \textbf{Average gesture surprise} - gesture surprise is calculated per-gesture as the
  elapsed proportional time since the previous gesture in the same localisation, or the start of the sample
  (proportional to length of the sample) for the first gesture.
  This overall feature averages the surprise value calculated for every gesture across all tracked localisations.
  I use the term \enquote{surprise} as the feature targets the effect on a gesture level basis, rather
  than the sample level.
  This is not a measure of how much of a sample no gesture is occurring as it is normalised on both
  the sample length and the number of gestures\footnote{
  To illustrate further: if 2 gestures occurred within a sample such that 80\% of the sample
  duration had no gesture occurring, the average gesture surprise would be $\dfrac{80}{2}=40$.
  Whereas, if there were 100 gestures, still with 80\% of the sample with no gesture
  occurring, the average surprise be 0.8\%, even though both samples had the same proportion without
  any gesture occurring.
  This matches the intuition that each gesture within 100 evenly spaced gestures would be unsurprising
  as they were regularly occurring, whereas the 2 evenly spaced gestures would be surprising because
  nothing was happening in between.
  }.
  \item \textbf{Average gesture movement standard deviation} - the standard deviation of per-frame
  movement within a gesture is averaged across all detected gestures.
  This is intended to indicate the consistency of movement intensity
  through a gesture.
  \item \textbf{Number of gestures} - total number of detected gestures across all tracked localisations.
\end{itemize}

\subsubsection*{Localised Features}
Whole body and localised features are concatenated in the same feature vector.
Localised features are included in the final feature vector \textbf{for each localisation included in the
vector}.

\begin{itemize}
  \item \textbf{Average length of gestures} - the average number of frames per gesture.
  \item \textbf{Number of gestures} - the total number of gestures, irrespective of gesture length.
  \item \textbf{Average per-frame gesture movement} - the average movement across all gestures.
  \item \textbf{Total movement in gestures} - the total amount of movement affected by the detected
  gestures.
  \item \textbf{Average gesture surprise} - the average surprise across all gestures.
\end{itemize}

\subsubsection*{Normalisation}
All features are normalised such that the length of the sample does not affect the results.

I normalise sum based features (e.g.\ gesture length, gesture count, total movement, etc) against
the total number of frames in the sample and against the total number of gestures for gesture average values.
For example, gesture surprise is normalised against the total number of frames and
normalised a second time against the total number of gestures.

\subsubsection*{Absent Features}
If a feature has no inputs (such as when no gesture was detected within a body localisation) its value is
set to $-1$ to enable models to incorporate the absence of movement in their predictions.

\subsection{Body Localisation}\label{subsec:pose-parts}
Gestures in different body localisations provide distinct information.
Aggregating gestures from different localisations provides a general representation of this information,
however, having features localised to specific body localisations provides further information, without
significantly increasing the dimensionality.

I define four localisations, hands, head, legs, and feet, based on specific pose estimation points.

\paragraph{Hands}
I use the finger tip points (including thumb) detected by OpenPose as the gesture detection points.
This means wrist based gestures (e.g.\ rolling of a hand) are detected.
Each hand is processed separately, that is, I detect gestures and calculate individual gestures
independently in each hand, these gestures are then aggregated into a single body localisation feature
vector.
This makes the final features side agnostic.
This ensures differences in dominant hand between participants will not affect the result.

\paragraph{Head}
While OpenPose includes face detection, providing detailed facial landmarks, I use the
general head position points provided by the standard pose detection component.

\paragraph{Legs}
I represent legs using the knee pose points from OpenPose.
As with hands, I process gestures in each leg independently and then aggregate to a single feature vector.

\paragraph{Feet}
Each foot is comprised of four pose points within OpenPose.
Aggregation is the same as hands and legs.

\paragraph{Trunk}
I do not include a \enquote{trunk} localisation as there is minimal movement in the trunk, given
the seated nature of the dataset interviews.
Though some participants may lean forwards and backwards, these movements are not represented well
within my data as the camera faces the participants directly such that forwards and backwards
leaning would be towards the camera, thus requiring depth perception which is not included in my
data.
Side to side leaning is restricted by the arms of the participant's chair.
As such the localisations that are relatively free to move, those other than the trunk, are
intuitively the most likely areas to provide predictive information.

\subsection{Gestures}\label{subsec:method-features-gestures}

\subsubsection*{Gesture Detection}\label{subsec:gesture-detection}

To detect gestures within a set of pose points (e.g.\ finger tips, knees, feet, head, etc) I scan the
activity of the target points for sustained periods of movement.
The gesture detection step takes cleaned per-frame pose estimations and outputs a collection of
ranges of non-overlapping frames that contain gestures within the localisation.

First, the per-frame absolute movement delta is calculated for each pose point.
The movement is then averaged across all localisation pose points.
Movement deltas are $L^2$ distances.
Formally,
\begin{equation}
  \begin{split}
    M_{p,t} & = P_{p,t} - P_{p,t-1} \\
    F_{t} & = \dfrac{1}{|P|}\sum_{p \in P} M_{p,t}
  \end{split}
  \label{eq:gesture-detection-1}
\end{equation}
where $M_{p, t}$ is the amount of movement for pose point $p$ at time $t$, $P_{p, t}$ is the position
value of pose point $p$ at time $t$, and $F_{t}$ is the averaged per-frame movement across all points.

Second, I average the movement of each frame within a window such that a small number of frames do not have a
disproportionate effect on the detection.
That is,
\begin{equation}
  \begin{split}
    W_{i} = \dfrac{1}{l}\sum_{t = i \times l}^{t < i \times (l + 1)}F_{t}
  \end{split}
\end{equation}
where $W_{i}$ is the windowed average at window index $i$, $l$ is the length of the window, and
$F_{t}$ is the average movement at frame $t$, from Equation~\ref{eq:gesture-detection-1}.
In this dissertation I use $l = 10$, i.e.\ a second of movement is represented by 3 windows, this is
experimentally chosen.

Third, the detector iterates through the averaged windows until a window with an average movement
above a threshold is found.
The first frame of this initial window is considered the beginning of the gesture.
The gesture continues until $n$ consecutive windows (I use 3, i.e.\ 30 frames, as an approximate
of a second) are found below the movement threshold.
The last frame of the final window above the movement threshold is considered the end of the gesture.
This is provided more formally in Algorithm~\ref{alg:gesture-detection}.

\begin{algorithm}
  \caption{Gesture detection}\label{alg:gesture-detection}
  \begin{algorithmic}[1]
    \State $n \gets \textit{number of consecutive windows}\text{ below threshold for end}$
    \State $m \gets \text{threshold of window }\textit{movement}$
    \State $l \gets \text{minimum }\textit{number of windows}\text{ for a gesture}$
    \State $gestures \gets EmptyList$
    \State $start \gets NULL$
    \State $belowThresholdCount \gets 0$

    \item[]
    \For{each window movement $W$ at index $i$}
      \If{$W \geq m$}
        \item[] \hskip\algorithmicindent\hskip\algorithmicindent
        \textit{// Start the gesture on the first window that exceeds the movement threshold.}
        \If{$start \equiv NULL$}
          \State $start \gets i$
        \EndIf
        \State $belowThresholdCount \gets 0$
        \item[]
      \ElsIf{$start \neq NULL$}
        \State $belowThresholdCount \gets belowThresholdCount + 1$
        \item[] \hskip\algorithmicindent\hskip\algorithmicindent
        \textit{// A gesture is completed after $n$ consecutive windows below the movement threshold.}
        \If{$belowThresholdCount \equiv n$}
          \item[] \hskip\algorithmicindent\hskip\algorithmicindent\hskip\algorithmicindent
          \textit{// The end of a gesture is the final window that exceeds the threshold.}
          \State $end \gets i - n$
          \If{$end - start \geq l$}
            \State $gestures$ append $[start,\ end]$
          \EndIf

          \item[] \hskip\algorithmicindent\hskip\algorithmicindent\hskip\algorithmicindent
          \textit{// Reset to find the next gesture.}
          \State $start \gets NULL$
          \State $belowThresholdCount \gets 0$
        \EndIf
      \EndIf
    \EndFor

    \item[] \item[] \item[] \textit{// Close the final gesture.}
    \If{$start \neq NULL$}
      \State $end \gets finalIndex$
      \If{$end - start \geq l$}
        \State $gestures$ append $[start,\ end]$
      \EndIf
    \EndIf

  \end{algorithmic}
\end{algorithm}

Having detected gestures in each body localisation I extract the features described above from each
gesture and aggregate them to form the final feature vector.

\section{Feature Space Search}\label{sec:method-feature-search}
I have described a collection of novel features whose individual and mutual predictive value is as yet
unknown.
Some features are potentially unhelpful to predictive models.
Therefore, distinguishing useful features from unhelpful features is a key operation to enable
development of accurate models, and thus validate these features.
To this end, I perform an exhaustive search of the feature space to identify the combination of features
with the best performance.

Alghowinem et al.\ \cite{Alghowinem:2015cs} demonstrate the benefits of a variable feature set,
achieving accuracy improvements of up to 10\%.
Their classification method is based on SVMs and the dimensionality reduction enabled by feature
filtering is significant.
They use a statistical T-threshold based approach to feature selection.
However, Dibeklioglu et al.\ \cite{Dibeklioglu:2015cf} argue that basing selection on optimization
of mutual information will achieve better results than individual information based selection.
I follow this mutual information approach and thus feature selection is based on the results achievable given
a combination of features, rather than features' individual relevance.

I define few enough features such that
a brute force feature combination space search (i.e.\ test every permutation) is viable.
As each feature can be included or excluded the space has $2^n$ permutations, where $n$ is the number
of features being searched.

I iterate over every permutation and perform three fold cross validation using the combination of
features, the permutation with the greatest average cross validation F1 score is taken as the best
permutation, which enables testing and evaluating the effectiveness of the proposed features.

\section{Summary}\label{sec:method-summary}

In this chapter I have described my methodology for extracting gesture meta features from videos.
The four core stages are: pose estimation, data preparation, feature extraction, and finally
classifier training.
I use OpenPose to extract pose data per-frame as it is the state-of-the-art in pose estimation.
I then filter samples and perform two operations to reduce noise within the remaining samples:
recovery of partial detection failures and smoothing of high frequency detection noise.
Features are extracted by detecting individual gestures, calculating per-gesture features such as
speed, and aggregating per-gesture features within their localisations and over the whole body.
Classifier choice is an evaluation detail and discussed in the next chapter.

\chapter{Evaluation}\label{ch:results}

Having described my gesture meta features, I now evaluate their predictive capability on the dataset
introduced in Chapter~\ref{ch:dataset}.
Section~\ref{sec:results-set-up} describes implementation details of my evaluations.
Section~\ref{sec:baseline} defines a simple baseline using a single feature.
In Section~\ref{sec:results-feature-representation} I evaluate the features' potential using the
exhaustive feature search described in Section~\ref{sec:method-feature-search}.
Section~\ref{sec:results-body-localisations} investigates the influence of different body localisations
on the features' effectiveness.
I then demonstrate the features' generalisability beyond the primary depression detection task in
Section~\ref{sec:results-generalisation}.
In Section~\ref{sec:results-comparison-method} I compare the features' automatic depression detection
performance with an existing body-modality method and a straightforward face-modality method.
In Section~\ref{sec:evaluation-multi-modal} I experiment with a multi-modal classifier
that combines the gesture meta features with face-modality features.
Finally, I summarize the results in Section~\ref{sec:evaluation-summary}.

\section{Implementation details}\label{sec:results-set-up}

Before presenting evaluation results, I outline the details of the evaluation setup.

\subsection*{Evaluation}\label{subsec:results-evaluation}
I perform evaluations using a three fold cross validation.
The training and test samples are split in a participant-independent manner and use stratified
folding to balance them with regards to labels.
Cross validating with more folds leads to fewer test samples per fold.
Given the small size of the dataset, this can lead to more erratic performance (i.e.\ more extremes
in cross validation results).
I assess results based on the average cross validation F1 score and the standard deviation between
fold F1 results.
Average F1 provides an objective measure of model quality.
Whilst the standard deviation provides an indicator of the consistency of the model.

\subsection*{Classifier Models}\label{subsec:results-models}
I evaluate four types of classifiers on all tasks:
\begin{itemize}
  \item A linear regression based classifier (denoted as \texttt{lin}) using a classification threshold of 0.5.
  \item A logistic regression based classifier (denoted as \texttt{log}) using the L-BFGS solver.
  \item A linear kernel SVM (denoted as \texttt{svm}) with balanced class weighting (without
  balancing the class weightings  the classifier consistently chooses to predict a single class on every fold).
  \item A random forest (denoted as \texttt{rf}) with 40 trees, feature bootstrapping, a minimum
  of 3 samples per leaf, a maximum depth of 5, balanced class weighting, and exposing 80\% of features
  per node.
  These parameters were chosen experimentally.
\end{itemize}

\subsection*{Labels}\label{subsec:results-setup-labels}
I only evaluate binary classification tasks (i.e.\ high vs.\ low depression score) within this dissertation.
However, the dataset contains continuous values for distress and personality measures,
thus a constant threshold is required for each label (similar to the participation selection criteria
discussed in Section~\ref{subsec:dataset-participants}).
These thresholds are chosen such that the resulting binary classes are as balanced as possible.
Given the small size of the dataset, balancing the classes is important to classifier training.
Per-label thresholds are reported in Table~\ref{table:label-thresholds}.

\begin{table}
  \small
  \begin{center}
    \begin{tabular}{rrrr}
      \toprule
      \textbf{Label} & \textbf{Threshold} & \textbf{\# Participants above} & \textbf{\# Participants below} \\ [0.5ex]

      \midrule
      \makecell[l]{\textbf{Distress}} &&& \\
      Depression & 7 & 18 & 17 \\
      Anxiety & 7 & 18 & 17 \\
      Perceived stress & 17 & 18 & 17 \\
      Somatic stress & 6 & 19 & 16 \\

      \midrule
      \makecell[l]{\textbf{Personality}} &&& \\
      Neuroticism & 17 & 18 & 17 \\
      Extraversion & 16 & 18 & 17 \\
      Agreeableness & 25 & 20 & 15 \\
      Conscientiousness & 20 & 20 & 15 \\
      Openness & 27 & 19 & 16 \\

      \bottomrule
    \end{tabular}
  \end{center}
  \caption{
  Binary classification thresholds for distress and personality labels.
  }
  \label{table:label-thresholds}
\end{table}

A larger dataset, or an expansion of this dataset, would enable regression models.
Regression models would also naturally provide a multi-class classification solution as the original
questionnaire scoring categories could be applied to the regressed predictions.

Though I evaluate the features' predictive capability against multiple labels, my primary focus is
on depression detection, as this is the area that most of the related work discusses.
An evaluation the features' capability on other labels is presented in Section~\ref{sec:results-generalisation}.

\subsection*{Localisations}
Though I evaluate four localisation types: head, hands, legs, and feet, the feet localisation
has a negative effect on performance, shown in Section~\ref{sec:results-body-localisations}.
As such, all other sections use the three useful localisations: head, hands, and legs, in their
evaluations.

\subsection*{Feature Reporting Notation}\label{subsec:results-feature-reporting-description}
To concisely report features used by different classification models I describe a brief notation for enumerating
features.
The notation defines tokens for localisations, feature type, and how the \texttt{lin} model
interprets the feature.
The structure is \enquote{\texttt{[localisation]-[feature type][linear polarity]}}.
Localisation and feature type token mappings are provided in Table~\ref{table:results-feature-notation-mapping}.

\begin{table}
  \begin{minipage}{.5\linewidth}
    \vspace*{-144pt}
    \begin{center}
      \begin{tabular}{cc}
        \toprule
        \textbf{Localisation} & \textbf{Token} \\ [0.5ex]
        \midrule
        Overall & \texttt{O} \\
        \cmidrule{1-2}
        Hands & \texttt{Hn} \\
        \cmidrule{1-2}
        Head & \texttt{He} \\
        \cmidrule{1-2}
        Legs & \texttt{L} \\
        \cmidrule{1-2}
        Feet & \texttt{F} \\
        \bottomrule
      \end{tabular}
    \end{center}
  \end{minipage}
  \begin{minipage}{.5\linewidth}
    \begin{center}
      \begin{tabular}{ccc}
        \toprule
        & \textbf{Feature} & \textbf{Token} \\ [0.5ex]
        \midrule
        \parbox[c]{3mm}{\multirow{5}{*}{\rotatebox[origin=c]{90}{\textbf{Overall\ \ \ \ \ \ \ \ \ \ \ \ }}}}
        & Average frame movement & \texttt{FM} \\ \cmidrule{2-3}
        & \makecell{Proportion of total movement \\ occurring during a gesture} & \texttt{GM} \\ \cmidrule{2-3}
        & Average gesture surprise & \texttt{GS} \\ \cmidrule{2-3}
        & \makecell{Average gesture movement \\ standard deviation} & \texttt{GD} \\ \cmidrule{2-3}
        & Number of gestures & \texttt{GC} \\
        \midrule
        \parbox[c]{3mm}{\multirow{5}{*}{\rotatebox[origin=c]{90}{\textbf{Localised\ \ \ \ \ \ }}}}
        & Average length of gesture & \texttt{GL} \\ \cmidrule{2-3}
        & \makecell{Average per-frame \\ gesture movement} & \texttt{GA} \\ \cmidrule{2-3}
        & Total movement in gestures & \texttt{GT} \\ \cmidrule{2-3}
        & Average gesture surprise & \texttt{GS} \\ \cmidrule{2-3}
        & Number of gestures & \texttt{GC} \\
        \bottomrule
      \end{tabular}
    \end{center}
  \end{minipage}
  \caption{Feature notation tokens.}
  \label{table:results-feature-notation-mapping}
\end{table}

I define an indicator of how a model interprets a feature based on its
contribution to a positive classification.
A greater value (e.g.\ more activity) contributing to a positive classification is denoted by \enquote{\texttt{+}}.
Conversely a greater value contributing to a negative classification is denoted by \enquote{\texttt{$\neg$}}.
A value which has a negligible effect (defined as a linear model applying a near-zero coefficient)
is denoted by \enquote{\texttt{/}}.
Finally, if the \texttt{lin} classifiers for each cross-validation fold are inconsistent in
usage of the feature the \enquote{\texttt{?}} indicator is used.

For example, \texttt{F-GS$\neg$} would denote that a greater amount of surprise in feet gestures
indicates a negative classification.

\section{Baseline}\label{sec:baseline}
As a baseline I evaluate the use of the only non-gesture based feature, average per frame movement
(\texttt{O-FM}).
Results are given in Table~\ref{table:baseline-f1}.

\begin{table}
  \begin{center}
    \begin{tabular}{rrrr}
      \toprule
      \textbf{Model} & \textbf{F1 avg} & \textbf{F1 std} \\ [0.5ex]

      \midrule
      \texttt{lin} & 34.43\% & 11.45\% \\
      \texttt{log} & 34.43\% & 11.45\% \\
      \texttt{svm} & 33.82\% & 10.73\% \\
      \texttt{rf} & \textbf{64.29\%} & \textbf{5.83\%} \\

      \bottomrule
    \end{tabular}
  \end{center}
  \caption{
  F1 aggregate scores for models using the baseline feature on the depression detection task.
  The \texttt{rf} model achieves the best performance.
  In this evaluation only one feature is used, this means that the feature is available to every decision node
  in the random forest, enabling strong overfitting.
  }
  \label{table:baseline-f1}
\end{table}

While the \texttt{rf} model achieves the best results, it is inconsistent across its folds.
Its 3 F1 scores are: 70.06\%, 60.00\%, and 47.06\%.
However, it does suggest that the movement feature alone is valuable for prediction.
Though, the worse than random results achieved by the other models suggest that movement is not a linear
indicator.
It is possible, indeed likely, that the \texttt{rf} model's result is reflective an overfitting of the dataset.

\section{Feature Search}\label{sec:results-feature-representation}
I evaluate the effect of the feature space search to identify the best feature combination.

\subsection*{All Features}
Table~\ref{table:results-feature-rep-baseline-all-features} presents results for each model when
provided with the full feature vector unmodified.
Given all features the \texttt{lin}, \texttt{log}, and \texttt{svm} models all improve on their baselines,
while the \texttt{rf} model is worse than its baseline.
The reduction in performance by the \texttt{rf} model can be attributed to the overfitting ability of
random forests.
This ability is especially prevalent with single feature problems as the feature must be made available
to every node within the forest, thus enabling direct fitting of multiple decision trees to the dataset.
The \texttt{lin} model achieves significantly better-than-random performance, indicating that the features
have some linear predictive capability.

\begin{table}
  \begin{center}
    \begin{tabular}{rrr}
      \toprule
      \textbf{Model} &
      \textbf{F1 avg} &
      \textbf{F1 std}
      \\ [0.5ex]

      \midrule
      \texttt{lin} & \textbf{66.81\%} & 8.89\% \\
      \texttt{log} & 56.55\% & \textbf{5.12\%} \\
      \texttt{svm} & 62.70\% & 9.19\% \\
      \texttt{rf} & 41.88\% & 6.04\% \\

      \bottomrule
    \end{tabular}
  \end{center}

  \caption{
  Classifier F1 performance for detecting depression within my dataset given all features.
  There are two primary interest points in this table: 1) the \texttt{lin} classifier performs best given
  all features, suggesting that the features provide some linear information, and 2) the \texttt{rf}
  classifier performs significantly worse than the one feature baseline in Table~\ref{table:baseline-f1},
  further suggesting the \texttt{rf} classifier is overfitting the one feature baseline.
  }
  \label{table:results-feature-rep-baseline-all-features}
\end{table}

\subsection*{Searched Features}
I perform an exhaustive feature space search to identify the best combination of features (determined
by the average cross-validation F1 score).
This provides two important outcomes: the best accuracy possible for a given model using these features
within my dataset and the most relevant features to predict depression within my dataset.
A good accuracy from the first outcome validates the features I have developed.
The second outcome then provides a basis for further analysis of these features and their relation
to depression within my dataset.

This search requires a model fast enough to train that the full space can be searched
in a viable amount of time and, preferably, a model that is interpretable.
For these reasons I use the \texttt{lin} model to search the feature space, and then evaluate the
best feature combination with the other classifier types.

\begin{figure}[t]
  \begin{center}
    \includegraphics[width=\linewidth]{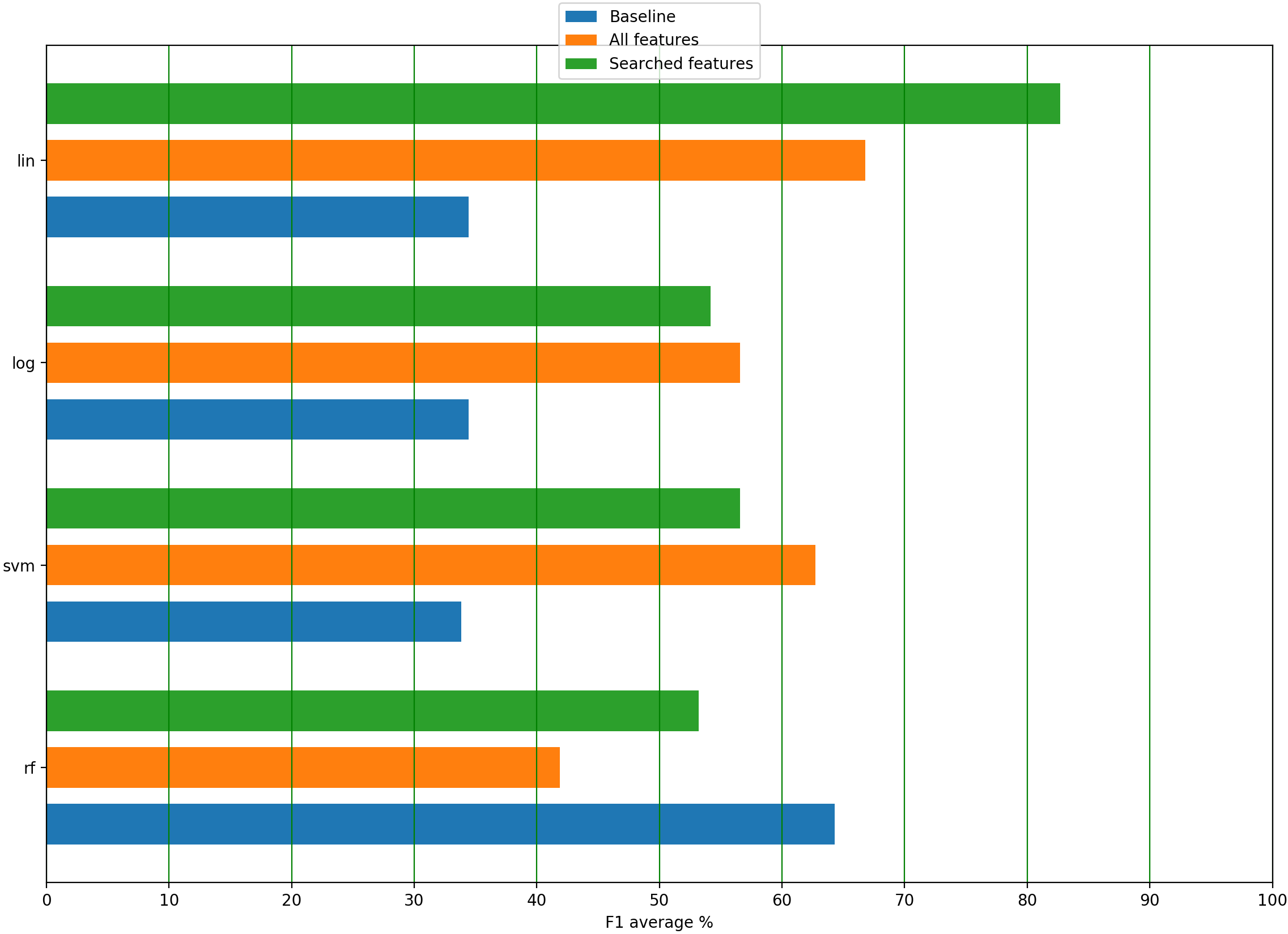}
  \end{center}
  \caption{
  Comparison of baseline results to feature searched results.
  This shows that performing a full feature search enables the \texttt{lin} classifier to outperform
  the other classifiers and feature combinations.
  Also of interest is the reduction in performance by the \texttt{rf} model when provided with more
  features, suggesting that it may be overfitting when provided with the single baseline feature.
  The specific feature combination resulting from the search is discussed below in \textbf{Chosen Features}.
  }
  \label{fig:baseline_comparison}
\end{figure}

\begin{table}
  \begin{center}
    \begin{tabular}{rrr}
      \toprule
      \textbf{Model} &
      \textbf{F1 avg} &
      \textbf{F1 std}
      \\ [0.5ex]

      \midrule
      \texttt{lin} & \textbf{82.70\%} & 8.95\% \\
      \texttt{log} & 54.17\% & 5.89\% \\
      \texttt{svm} & 56.55\% & \textbf{5.12\%} \\
      \texttt{rf} & 53.18\% & 14.96\% \\

      \bottomrule
    \end{tabular}
  \end{center}
  \caption{
  Performance of the best feature combination as determined by an exhaustive feature space search
  using the \texttt{lin} classifier to detect depression.
  This demonstrates the power, and the linearity, of the gesture meta features as the \texttt{lin}
  classifier is able to achieve a high F1 score.
  }
  \label{table:results-feature-search}
\end{table}

\subsubsection*{Feature Search Improves Best Performance}
The best performance when applying the feature search, again from the \texttt{lin} classifier, is
significantly better, 82.70\%, than the classifier's baseline of 34.43\% and all features baseline
of 66.81\%.
This demonstrates the importance of reducing dimensionality and feature confusion, especially when
using relatively direct methods such as linear regression.
Results are provided in Table~\ref{table:results-feature-search}, a comparison of these results to the
baseline results is presented in Figure~\ref{fig:baseline_comparison}.

\subsubsection*{Chosen Features}
The full feature combination is:
\{\texttt{O-FM?, O-GM+, O-GC?, Hn-GC?, Hn-GT$\neg$, Hn-GS?, He-GL?, He-GC?, He-GA+, He-GT$\neg$, He-GS?, L-GL$\neg$, L-GC?, L-GT+}\}.
Analysing this feature set, we can derive some intuition as to information indicative of depression.
The overall (\texttt{O-*}) features suggest that the number of gestures (\texttt{O-GC}) and the
amount of movement within those gestures (\texttt{O-GM}) is relevant to depression.
The \texttt{O-GM+} token suggests that more movement within gestures relative to all other movement
is indicative of depression.
The localised features suggest that the length of gestures (\texttt{*-GL}) has a correlation with depression,
however, this correlation differs between localisations.
The \texttt{head} localisation is ambiguous as to whether shorter or longer gestures (\texttt{He-GL?}) is
indicative of depression.
Whilst longer gestures in the legs localisation (\texttt{L-GL$\neg$}) is indicative of less depression.
Within this model, less total movement of the hands (\texttt{Hn-GT$\neg$}) is indicative of distress.

\subsubsection*{Negative Performance Impact on Other Models and Overfitting}
The identified feature set is chosen using the \texttt{lin} model, so it is unsurprising it has
a greater improvement than any other model.
While the \texttt{log} classifier's performance does not change much, the \texttt{svm} and \texttt{rf}
classifiers have reduced performance compared to their all features and one feature baselines, respectively.
There are two aspects to consider here: the value of pre-model filtering to each model and the potential
for each model to overfit.
Focusing on the \texttt{rf} classifier as it has a more distinct reduction in performance;
random forests have inbuilt feature discretion, so the pre-model filtering of features does not have
as great a complexity reducing effect as it does on the other models.
My hyper-parameters give the \texttt{rf} model a relatively large selection of features (80\%) per
node, thus it should generally filter, to some extent, those naturally unhelpful features.
Random forests, as decision tree ensemble methods, have a propensity for overfitting data.
By reducing the number of features available I reduce the available surface for overfitting.
Moreover, when only using one feature, as in the baseline, every decision node in the random forest
has access to the feature, thus enabling particularly strong overfitting of a small dataset.

\subsection*{Dimensionality Compression}
I do not perform any dimensionality compression operations in my presented evaluations.
However, I have experimented with Principle Component Analysis (PCA) both independently and in
combination with a feature search.
Neither approach achieved especially interesting results, all were worse than when not applying PCA\@.
Given this, and the already relatively low number of dimensions, I do not see it as a critical path
of investigation for this dissertation.

\section{Body Localisations}\label{sec:results-body-localisations}
Not all localisations are necessarily beneficial.
Identifying which localisations are helpful is made more difficult by the localisations' interactions
within a classifier and the effect they have on the overall features.
Though a localisation may generate features that are chosen using feature search, they may reduce
overall accuracy by obfuscating predictive information in the overall features.

\begin{figure}[t]
  \begin{center}
    \includegraphics[width=\linewidth]{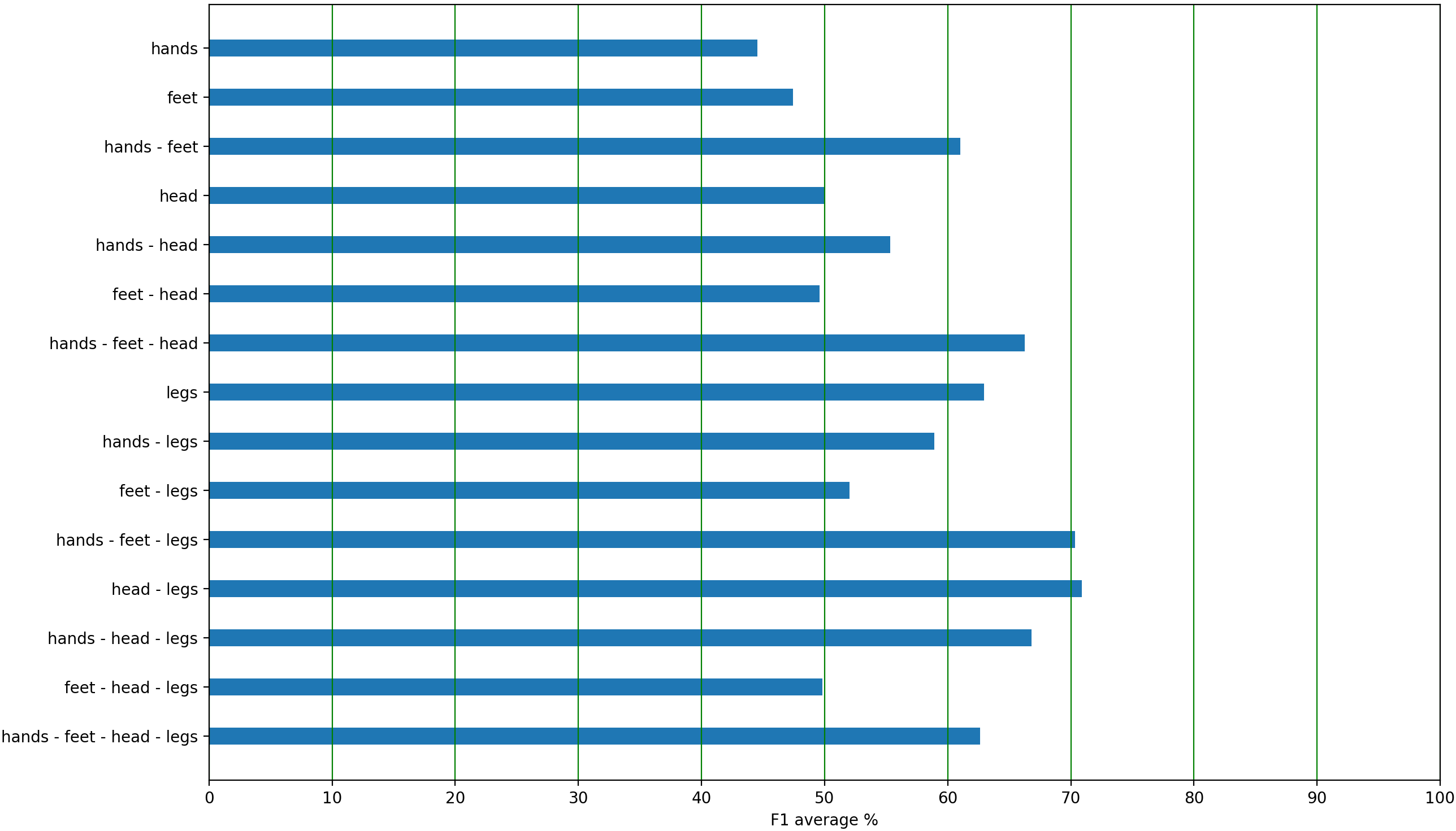}
  \end{center}
  \caption{
  Comparison of the best F1 average scores from localisation combinations using the \texttt{lin} classifier
  using all features.
  The most interesting results are the bottom four (vertically) localisation results:
  \texttt{head - legs}, \texttt{hands - head - legs}, \texttt{feet - head - legs}, and
  \texttt{hands - feet - head - legs}.
  Specifically, the feet localisation impairs the performance of the localisation combinations
  when included.
  This trend is also seen in \texttt{feet - legs} and \texttt{feet - head}.
  }
  \label{fig:localisation_comparison}
\end{figure}

Given this, I experiment with localisations included individually and in varying combinations.
I also provide an example of a localisation, feet, that negatively effects performance when included,
even when all other localisations are also included (and are thus providing the same predictive
information).
A comparison of the best F1 scores for localisation combinations is presented in
Figure~\ref{fig:localisation_comparison}.

\subsubsection*{Localisation Inclusion}
Clearly not all of the features generated per localisation provide useful information.
Inclusion of more localisations, and thus a larger feature space, does not guarantee a better
optimal result is available within the space.
As the overall features (those aggregated across all localisations) are effected by each localisation,
the quality of the feature space can be degraded with the inclusion of localisations.
For example, this occurs regularly in Figure~\ref{fig:localisation_comparison} when including
the \texttt{feet} localisation.
In particular, the best base configuration, \texttt{head - legs} using the \texttt{lin} classifier,
achieves a 70.88\% F1 score, when the \texttt{feet} localisation is included this drops to 49.84\%.
I have not identified any psychology literature, or clear intuition, as to why the \texttt{feet}
localisation hinders performance.
I see three probable explanations: 1) some literature does support this and I have simply
not identified the literature, 2) this is accurately representing that feet movement meta information does not
distinctively change with distress, but no literature has explicitly investigated this, and 3) this
is simply a attribute of the dataset that is not
reflective of any broader trend, either due to the dataset size or the nature of the interview dynamic.

\subsubsection*{Best Base Performance Configuration}
Though the \texttt{head - legs} configuration achieves the best performance when all features are
used, it does not achieve the best performance when features are chosen based on an exhaustive search.
While my primary configuration, \texttt{hands - head - legs}, achieves 82.70\% F1 average, the
\texttt{head - legs} configuration achieves 80.53\%, results are presented in
Table~\ref{table:results-localisations-head-legs}.

\begin{table}
  \begin{center}
    \begin{tabular}{rrr}
      \toprule
      \textbf{Model} & \textbf{F1 avg} & \textbf{F1 std} \\ [0.5ex]

      \midrule
      \texttt{lin} & \textbf{80.53\%} & \textbf{4.04\%} \\
      \texttt{log} & 59.52\% & 8.91\% \\
      \texttt{svm} & 65.40\% & 8.40\% \\
      \texttt{rf} & 51.32\% & 5.76\% \\

      \bottomrule
    \end{tabular}
  \end{center}
  \caption{
  Performance of models when using features chosen via exhaustive search with source features from
  the \texttt{head - legs} configuration.
  This configuration achieves close to the best performance of the standard configuration
  (Table~\ref{table:results-feature-search}).
  The \texttt{lin} classifier also has more consistent performance across cross validation folds than
  it does on the standard configuration, with a standard deviation of 4.04\% compared to 8.95\%.
  However, these results do not clearly define which configuration is generally better as the
  differences are quite minor.
  }
  \label{table:results-localisations-head-legs}
\end{table}

\section{Generalisability}\label{sec:results-generalisation}
I have demonstrated the gesture meta features' predictive value with regards to depression detection.
I now evaluate their predictive capability for other labels including participants' gender,
personality measures, anxiety, perceived stress, and somatic stress.

\begin{figure}
  \begin{center}
    \includegraphics[width=\linewidth]{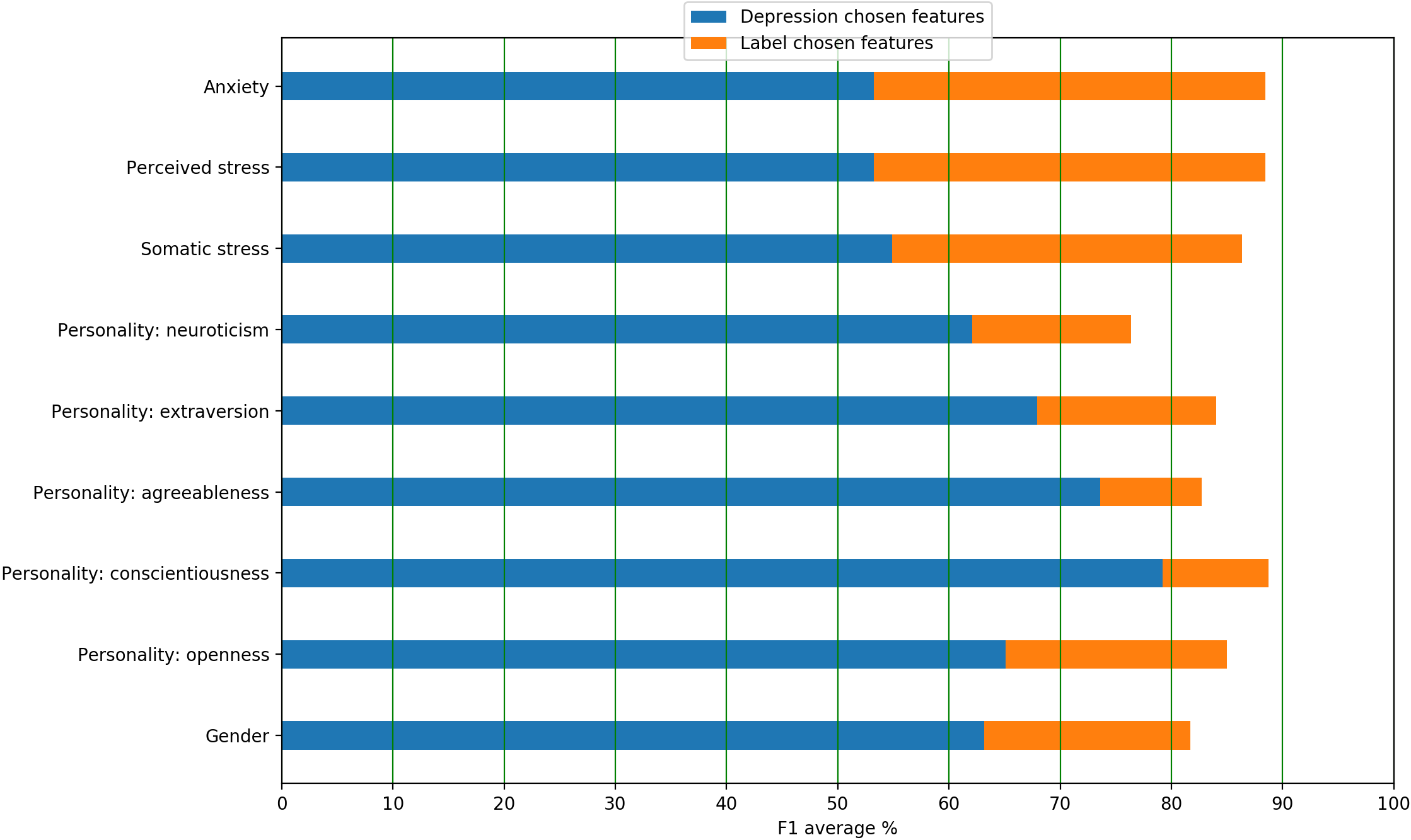}
  \end{center}
  \caption{
  Comparison of F1 average scores when using the optimal feature combination for the depression label
  vs.\ the optimal feature combination for each label.
  Each label has a significant performance improvement when using its optimal feature combination.
  }
  \label{fig:label_feature_comparison}
\end{figure}

I apply the best feature combination identified in Section~\ref{sec:results-feature-representation}
to each of the labels, presented in Table~\ref{table:results-other-labels-standard-features}.
I also perform a feature space search for each label, using the same process as
Section~\ref{sec:results-feature-representation}, to provide greater insight into the effect of features
and their reliability across labels, presented in Table~\ref{table:results-other-labels-searched-features}.
A comparison is shown in Figure~\ref{fig:label_feature_comparison}.
Consistent identification of features across uncorrelated labels reinforces the hypothesis that they provide
predictive information beyond a specific label and dataset (i.e.\ are less likely to be overfitting).

\begin{table}
  \tiny
  \begin{center}
    \begin{tabular}{rrrrrrrrr}
      \toprule
      \textbf{Label} &
      \multicolumn{2}{c}{\texttt{lin}} &
      \multicolumn{2}{c}{\texttt{log}} &
      \multicolumn{2}{c}{\texttt{svm}} &
      \multicolumn{2}{c}{\texttt{rf}}
      \\ [0.5ex]
      \midrule
      &
      \textbf{F1 avg} & \textbf{F1 std} & % lin
      \textbf{F1 avg} & \textbf{F1 std} & % log
      \textbf{F1 avg} & \textbf{F1 std} & % svm
      \textbf{F1 avg} & \textbf{F1 std} \\ % rf

      \midrule
      \makecell[l]{\textbf{Distress}} &&&&&&&  \\
      Depression & \textbf{82.70\%} & 8.95\% & 54.17\% & 5.89\% & 56.55\% & \textbf{5.12\%} & 53.18\% & 14.96\% \\
      Anxiety & 47.18\% & 23.21\% & 38.33\% & 19.20\% & 30.18\% & \textbf{4.02\%} & \textbf{53.26\%} & 16.58\% \\
      Perceived stress & 47.18\% & 23.21\% & 38.33\% & 19.20\% & 30.18\% & \textbf{4.02\%} & \textbf{53.26\%} & 16.58\% \\
      Somatic stress & 42.74\% & 30.29\% & 51.68\% & \textbf{4.01\%} & 44.44\% & 6.29\% & \textbf{54.89\%} & 8.37\% \\
      \midrule
      \makecell[l]{\textbf{Personality}} &&&&&&&  \\
      Neuroticism & \textbf{62.08\%} & \textbf{0.76\%} & 31.71\% & 10.89\% & 33.36\% & 12.59\% & 38.30\% & 8.30\% \\
      Extraversion & 51.04\% & 14.06\% & \textbf{67.95\%} & 17.33\% & 65.14\% & \textbf{12.98\%} & 52.94\% & 20.83\% \\
      Agreeableness & 69.48\% & 5.97\% & \textbf{73.61\%} & \textbf{2.13\%} & 67.98\% & 3.92\% & 56.96\% & 20.79\% \\
      Conscientiousness & 71.28\% & 6.53\% & 72.95\% & 6.93\% & 78.77\% & 6.24\% & \textbf{79.19\%} & \textbf{6.11\%} \\
      Openness & 49.64\% & 14.94\% & \textbf{65.08\%} & 5.94\% & 64.46\% & \textbf{5.31\%} & 61.47\% & 8.64\% \\
      \midrule
      \makecell[l]{\textbf{Demographic}} &&&&&&&  \\
      Gender & 34.26\% & 18.18\% & 52.96\% & \textbf{2.28\%} & 40.63\% & 18.42\% & \textbf{63.14\%} & 5.29\% \\

      \bottomrule
    \end{tabular}
  \end{center}
  \caption{
  Performance of models using the feature combination identified for the depression label (Section~\ref{sec:results-feature-representation})
  for predicting a variety of labels.
  The best results per-label are \textbf{bolded}.
  These results suggest that the depression chosen feature combination does not generalise particularly well.
  These results are also surprising as the distress labels (anxiety, perceived stress,
  and somatic stress), which are correlated to the depression label, perform poorly, whilst uncorrelated
  labels (such as agreeableness, openness, and gender) perform better.
  This may be due to overfitting of feature profiles to labels (i.e.\ the optimal features for the label)
  or truly distinct feature profiles between the correlated distress labels, though the former appears
  more probable.
  }
  \label{table:results-other-labels-standard-features}
\end{table}

\begin{table}
  \tiny
  \begin{center}
    \begin{tabular}{rrrrrrrrr}
      \toprule
      \textbf{Label} &
      \multicolumn{2}{c}{\texttt{lin}} &
      \multicolumn{2}{c}{\texttt{log}} &
      \multicolumn{2}{c}{\texttt{svm}} &
      \multicolumn{2}{c}{\texttt{rf}}
      \\ [0.5ex]
      \midrule
      &
      \textbf{F1 avg} & \textbf{F1 std} & % lin
      \textbf{F1 avg} & \textbf{F1 std} & % log
      \textbf{F1 avg} & \textbf{F1 std} & % svm
      \textbf{F1 avg} & \textbf{F1 std} \\ % rf

      \midrule
      \makecell[l]{\textbf{Distress}} &&&&&&&  \\
      Depression & \textbf{82.70\%} & 8.95\% & 54.17\% & 5.89\% & 56.55\% & \textbf{5.12\%} & 53.18\% & 14.96\% \\
      Anxiety & \textbf{88.46\%} & 9.53\% & 46.14\% & \textbf{9.16\%} & 54.33\% & 12.32\% & 52.94\% & 12.85\% \\
      Perceived stress & \textbf{88.46\%} & 9.53\% & 46.14\% & \textbf{9.16\%} & 54.33\% & 12.32\% & 52.94\% & 12.85\% \\
      Somatic stress & \textbf{86.37\%} & \textbf{6.45\%} & 58.44\% & 8.85\% & 68.08\% & 16.97\% & 49.48\% & 11.38\% \\
      \midrule
      \makecell[l]{\textbf{Personality}} &&&&&&&  \\
      Neuroticism & \textbf{76.39\%} & 8.56\% & 38.85\% & 5.15\% & 48.25\% & \textbf{4.47\%} & 40.78\% & 13.59\% \\
      Extraversion & \textbf{84.04\%} & \textbf{10.24\%} & 74.61\% & 10.34\% & 73.39\% & 12.75\% & 58.59\% & 31.33\% \\
      Agreeableness & \textbf{82.70\%} & 5.77\% & 69.14\% & 6.67\% & 50.43\% & 14.85\% & 67.97\% & \textbf{1.85\%} \\
      Conscientiousness & \textbf{88.72\%} & 4.25\% & 73.14\% & 3.13\% & 69.90\% & 7.84\% & 78.91\% & \textbf{2.80\%} \\
      Openness & \textbf{85.01\%} & 6.62\% & 64.32\% & \textbf{5.06\%} & 68.77\% & 8.84\% & 60.32\% & 7.36\% \\
      \midrule
      \makecell[l]{\textbf{Demographic}} &&&&&&&  \\
      Gender & \textbf{81.69\%} & 5.32\% & 68.11\% & 10.22\% & 76.71\% & \textbf{5.10\%} & 64.81\% & 11.42\% \\

      \bottomrule
    \end{tabular}
  \end{center}

  \caption{
  Performance of models for predicting a variety of labels
  when using features identified via a feature space search specific to the label.
  This demonstrates the performance improvement, compared to Table~\ref{table:results-other-labels-standard-features},
  achieved via label specific feature combinations.
  The best results for each label are \textbf{bolded}.
  }
  \label{table:results-other-labels-searched-features}
\end{table}

\subsection*{Results}
The depression feature combination does not generalise well to the other distress measures for the \texttt{lin}
model, though it achieves 62--71\% F1 scores for neuroticism, agreeableness, and conscientiousness.
Interestingly, the other labels' best results using the depression combination are above 60\% F1, with the
exception of the other distress measures, which only achieve a best of 53--54\% F1.
This is surprising as the distress labels are strongly correlated to the depression label.

However, when features are chosen on a per-label basis the results improve significantly.
All labels\footnote{
Though perceived stress and anxiety measures have a covariance of 82.98\% within my dataset, once reduced
to binary classifications they are equivalent (i.e.\ 100\% correlation).
Thus results for both labels are the same.
} achieve 76\%+ (all but one are above 80\%) average F1 score with their best classifier.
The best classifier for all labels is the \texttt{lin} classifier, which is to be expected given
it is the search classifier and previous evaluation has shown the features provide useful linear
information.

\subsubsection*{Fitting features to labels}
There are two potential explanations for the substantial improvement in performance when using
label specific feature combinations: each label
could legitimately have a unique feature profile through which its distinctive attributes are
expressed or the labels could be experiencing a level of overfitting due to the small size of the
dataset.
While it is likely to be a combination of the two reasons, labels that are relatively uncorrelated
with depression are still able to achieve good performance using the depression chosen features,
suggesting it is not just overfitting.
For example, agreeableness, extraversion, and conscientiousness all have covariances of less than
50\% with depression, yet they achieve 72.61\%, 67.95\%, and 79.19\% average F1 scores, respectively,
with the depression features.
Openness which has a 4.29\% covariance with depression achieves 65.08\% with the depression
chosen features.
Furthermore, as Table~\ref{table:eval-gen-features} shows, the openness feature combination
shares only 6 out of its 9 features with the depression combination, whilst it also excludes 8
out of 14 of the depression combination's features.
Therefore, it can be reasonably suggested that the features do generalise, acknowledging
that fitting the features directly to a label achieves the best performance, as would be expected
with all prediction tasks.

\subsubsection*{Cross classifier generalisability}
Within the depression feature combination results, the \texttt{lin}, \texttt{log}, and \texttt{rf} models
all achieve the best results on at least 2 labels each.
The \texttt{svm} classifier performs close to the best on many labels, such as conscientiousness
where it achieves 78.77\% compared to the best result of 79.19\%.

The \texttt{svm} model performs better when classifying conscientiousness, extraversion, and
agreeableness, using the depression feature combination, than it does when predicting the depression label.
It is also better at transferring the depression feature combination to other labels than the \texttt{lin}
model that the feature combination was chosen with.
Its performance improves on most labels when using the label targeted feature sets, such as on
gender where it achieves 76.71\% and somatic stress with 68.08\% F1, whilst with the depression
feature combination its F1 result was less than 50\% for both.
Interestingly, whilst the \texttt{svm} model performed particularly well on agreeableness using
the depression feature combination, it performed worse when using the label targeted feature combination.

\subsubsection*{Personality}
Within the results for the depression feature combination the conscientiousness label achieves the best average
F1 score across all classifiers, 75.55\%, with an average standard deviation of 6.45\%.

Unlike the distress evaluation questionnaires, the BFI (personality) questionnaire is designed such
that responses should remain relatively consistent over time.
Future work could investigate whether the features are able to consistently predict personality measures
over time for a participant.
Do the features remain predictive of personality as a person's temporal distress measures change?

\subsubsection*{Chosen Features}
Table~\ref{table:eval-gen-features} presents the feature sets resulting from the per-label
feature searches.
There are some consistencies and intuitive inversions within the feature sets.
For example, the head localised average gesture movement (\texttt{He-GA}) is relevant for
many of the labels, though in some it is inconsistent whether greater or lesser is indicative of the
label (instances of inconsistencies are denoted as \texttt{He-GA?} within
Table~\ref{table:eval-gen-features}).
The feature usage is inverted between neuroticism, where a faster average speed is indicative of a positive
classification, and extraversion and agreeableness, where a slower average speed is indicative
of positive classification.

Furthermore, as stated above, features that are chosen by uncorrelated labels support the hypothesis
that these features are providing useful information.
For example, of the 7 features the gender label chooses, 5 are also chosen by the anxiety label,
including \texttt{hand} total gesture movement (\texttt{Hn-GT}), \texttt{hand} gesture surprise
(\texttt{Hn-GS}), and \texttt{leg} gesture surprise (\texttt{L-GS}).
Whilst gender and anxiety have a covariance of 7.23\% within the dataset.

\begin{table}
  \tiny
  \begin{center}
    \begin{tabular}{rrrrrr}
      \toprule
      \textbf{Label} &
      \textbf{\# Feat.} &
      \multicolumn{4}{c}{\textbf{Localizations}}
      \\ [0.5ex]

      \midrule
      & &
      \multicolumn{1}{c}{\textbf{Overall}} &
      \multicolumn{1}{c}{\textbf{Hands}} &
      \multicolumn{1}{c}{\textbf{Head}} &
      \multicolumn{1}{c}{\textbf{Legs}} \\

      \midrule
      \makecell[l]{\textbf{Distress}} &&&&  \\
      Depression & 14 & \texttt{FM?, GM+, GC?} & \texttt{GC?, GT$\neg$, GS?} & \texttt{GL?, GC?, GA+, GT$\neg$, GS?} & \texttt{GL$\neg$, GC?, GT+} \\
      Anxiety & 13 & \texttt{FM+, GM+, GD$\neg$, GC$\neg$} & \texttt{GT?, GS+} & \texttt{GL+, GA?, GT$\neg$} & \texttt{GL$\neg$, GA+, GT+, GS+} \\
      Perceived stress & 13 & \texttt{FM+, GM+, GD$\neg$, GC$\neg$} & \texttt{GT?, GS+} & \texttt{GL+, GA?, GT$\neg$} & \texttt{GL$\neg$, GA+, GT+, GS+} \\
      Somatic stress & 9 & \texttt{GD$\neg$} & \texttt{GL?, GT?, GS+} & \texttt{GC$\neg$, GA+} & \texttt{GL$\neg$, GC$\neg$, GT+} \\
      \midrule
      \makecell[l]{\textbf{Personality}} &&&&  \\
      Neuroticism & 14 & \texttt{GM+, GS?, GC$\neg$} & \texttt{GC+, GA$\neg$} & \texttt{GL?, GC+, GA+, GT$\neg$} & \texttt{GL$\neg$, GC+, GA?, GT+, GS+} \\
      Extraversion & 10 & \texttt{FM+, GC+} & \texttt{GL$\neg$, GC+, GA+} & \texttt{GC?, GA$\neg$} & \texttt{GL+, GT$\neg$, GS$\neg$} \\
      Agreeableness & 9 & \texttt{GS+, GC+} & \texttt{GC$\neg$, GA$\neg$} & \texttt{GA$\neg$, GS+} & \texttt{GC$\neg$, GA?, GT?} \\
      Conscientiousness & 8 & \texttt{FM?, GM$\neg$, GS?} & \texttt{GL+, GS?} & \texttt{GC+} & \texttt{GL?, GC?} \\
      Openness & 9 & \texttt{GM+, GD+} & \texttt{GL$\neg$, GC?} & \texttt{GC$\neg$, GA?, GS?} & \texttt{GA?, GT?} \\
      \midrule
      \makecell[l]{\textbf{Demographic}} &&&&  \\
      Gender & 7 & \texttt{GS+, GD$\neg$} & \texttt{GC+, GT+, GS$\neg$} & \texttt{GA+} & \texttt{GS$\neg$} \\

      \bottomrule
    \end{tabular}
  \end{center}

  \caption{
  Features chosen for each label when features are searched specifically for the label.
  \enquote{Positive classification} for the gender label is female (i.e.\ \texttt{He-GA+} indicates
  a higher average speed of head gestures is indicative of the participant being female).
  Refer to Table~\ref{table:results-feature-notation-mapping} for the full notation mapping.
  }
  \label{table:eval-gen-features}
\end{table}

\subsubsection*{Movement}
I represent four measures of movement: average movement per-frame (i.e.\ speed), standard deviation
of movement (i.e.\ consistency of speed), total movement both over an entire sample and over individual
gestures, and proportion of total movement that occurs during gestures (i.e.\ how much/little movement
is there outside of gestures).
The amount and speed of movement is intuitively correlated to distress and personality.
This intuition is validated as these movement representations are included in a variety of feature
sets resulting from feature space searches, across multiple labels.
Indeed, Table~\ref{table:eval-gen-features} shows that the speed of the head during
a gesture (i.e.\ \texttt{He-GA}) is correlated with positive classifications for labels including
neuroticism, depression, somatic stress, while it is inversely correlated with other classifications
including extraversion and agreeableness.

\subsubsection*{Surprise}
One of the less initially intuitive features is \enquote{gesture surprise}.
This feature represents the distance between a gesture and the previous gesture (or beginning of
the sample) as a proportion of the total length of the sample.
This per-gesture value is then averaged across all gestures, this means it is not a measure of the
proportion of the sample when no gesture occurs (as discussed in Section~\ref{sec:feature-extraction}).
The intuition is to represent whether the participant's gestures occur regularly or after a period
of stillness (i.e.\ how \enquote{surprising} is the gesture).
Gesture surprise is included in every feature combination resulting from a feature search of a label,
shown in Table~\ref{table:eval-gen-features}, suggesting it provides useful information.

\subsubsection*{Linearity}
The features are, in general, linearly correlated with different labels, as shown by the effectiveness
of the \texttt{lin} classifier, which has been the focus of the feature development and evaluation.
Using this classifier provides interpretability of the features and integrity that the features are
providing real, useful, information, not simply a platform for overfitting.

However, not all features are consistently linear with regards to certain labels.
Some features are learnt as correlated and inversely correlated on the same label in different
cross validation rounds.
This inconsistency denoted with the usage indicator \enquote{\texttt{?}}.
For example, while the \texttt{hands} localised gesture surprise feature (\texttt{Hn-GS}) is linearly
correlated with somatic stress, anxiety, perceived stress, and gender, it is inconsistently correlated
with conscientiousness and depression.
These inconsistencies within feature correlation are the exception, not the rule.
Indeed, all of the features that experience inconsistency on a label are shown to be linearly correlated
to another label, or linearly correlated within another localisation.

\section{Comparison with Related Work}\label{sec:results-comparison-method}
I evaluate two comparison methods: a linear SVM model based on Bag-of-Words features of FACS~\cite{Ekman:Rzx-k0OT} Action
Units (AUs), and the Bag-of-Body Dynamics (BoB) method presented by Joshi et al.\ \cite{Joshi:uf}.
The first method provides a comparison of modality (facial-modality) within my dataset and a cross-dataset
benchmark with the closest related dataset, DAIC\@.
The second method provides a comparison to an existing body-modality method that uses an
automatic expression categorization approach to predict depression.

\subsection*{AU SVM}
This method uses basic facial expression analysis to predict a binary depression label.
I implement a simple linear kernel SVM based on a Bag-of-Words (BoW) feature vector of FACS AUs to predict the
binary depression label in my dataset and, separately, a binary depression label within DAIC\@.
As with my dataset, I apply a threshold to DAIC's PHQ-8 score labels such that the dataset is as
balanced as possible.
The resulting threshold is 5 (i.e.\ 6 and above is considered \enquote{depressed}), this results in 68 of 139
samples classed as \enquote{depressed}.
This is a lower threshold than used in my dataset (which is 7).

Within the BoW feature vector, binary AUs are counted for each frame they are detected in, whilst
intensity measured AUs (i.e.\ continuous value measures) are summed across all frames.
The sums and counts are then normalised against the number of frames in which the face was successfully
detected.

The DAIC dataset provides per-frame predictions for 20 AUs.
I use OpenFace~\cite{Baltrusaitis:bv} to extract per-frame AUs from my dataset.
OpenFace predicts 35 AUs per-frame, substantially more than provided by DAIC\@.
While this method does not use the gesture features, I still only evaluate it on the samples that pass
the gesture filtering step (Section~\ref{subsec:filtering}) so that the evaluation is consistent
with previous results.

\subsection*{Bag-of-Body Dynamics}
Joshi et al.\ \cite{Joshi:uf} use a BoW approach to predict depression in clinically assessed
participants from their BlackDog dataset.
The BoW feature vector is comprised of categorized body expressions based on K-Means clustering
of histograms surrounding STIPs within a video of the subject.
Their method is:
\begin{enumerate}
  \item Extract Space-Time Interest Points (STIPs).
  \item Calculate histograms of gradients (HoG) and optic flow (HoF) around the STIPs.
  \item Cluster the histograms per-sample using K-Means, the resulting cluster centres define the
  sample's \enquote{key interest points} (KIPs).
  \item Cluster KIPs across all samples to define a BoB codebook, also using K-Means.
  \item Generate a BoB feature vector for each sample by fitting its KIPs to the codebook.
  \item Finally, apply a non-linear SVM to the resulting feature vectors to predict depression.
\end{enumerate}
They use a radial basis function kernel (RBF) for their SVM\@.

\paragraph{Differences in Method}
In the original paper Joshi et al.\ experiment with multiple KIP counts, codebook sizes, and
perform an extensive grid search for their SVM parameters.
I use a single KIP count and codebook size, 1,000 and 500, which they also experiment with.
While I perform a grid search on the RBF parameters, it may not be as extensive as their search.

I also exclude samples that generate fewer STIPs than the KIP count.
This results in 45 out of 65 samples being \textbf{included}.

\subsection*{Results}
\begin{figure}[t]
  \begin{center}
    \includegraphics[width=0.9\textwidth]{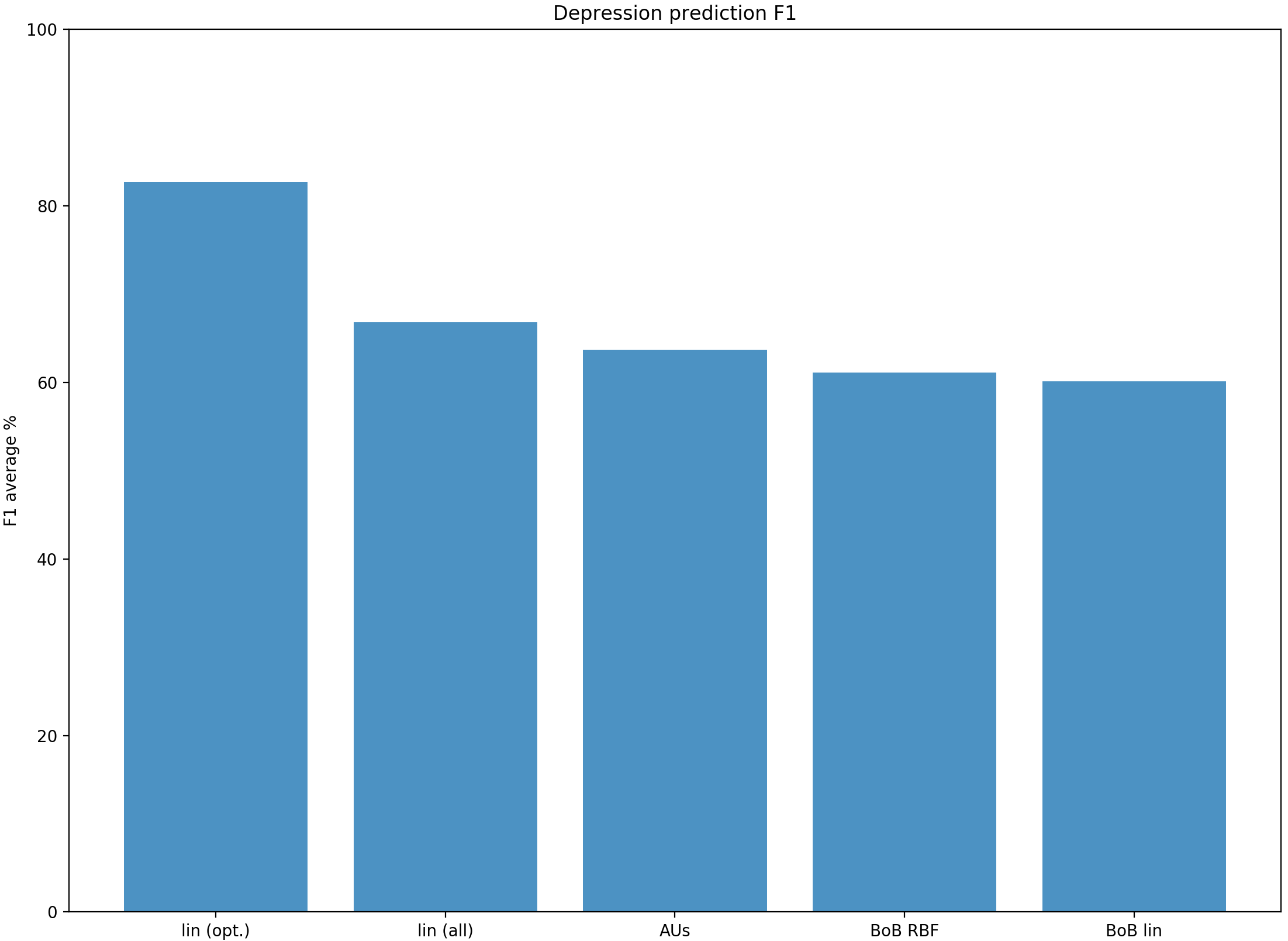}
  \end{center}
  \caption{
  Comparison of method F1 scores for predicting the depression label in my dataset.
  This shows the \texttt{lin} classifier using the optimal depression feature combination identified
  in Section~\ref{sec:results-feature-representation} (\texttt{lin (opt.\@)}) outperforming the comparison methods.
  The \texttt{lin} classifier using all of the gesture meta features (\texttt{lin (all)}) also
  outperforms the comparison methods, though not by as much.
  There are two caveats to this comparison: the features for the \texttt{lin (opt.\@)} classifier
  have been fitted to the label and the BoB results are for a reduced dataset (45 samples compared to 53
  for the \texttt{lin} and AUs models) as not all of the samples produced enough STIPs for the
  methods to operate appropriately.
  }
  \label{fig:comparison_methods}
\end{figure}

The results for each model are presented in Table~\ref{table:results-comparison-of-methods}, a
comparison of methods applied to my dataset is shown in Figure~\ref{fig:comparison_methods}.
I compare the best depression detection model from my previous evaluations (i.e.\ the feature combination
identified in Section~\ref{sec:results-feature-representation} with the \texttt{lin} classifier),
and a \texttt{lin} classifier using all gesture meta features, with
two BoB based SVMs and a FACS AUs based linear SVM for predicting depression.
The \texttt{lin} model performs best, achieving an 82.70\% F1 average with the optimal feature combination
and 66.81\% with all features, compared to the next best at
63.71\% from the FACS AUs SVM model and 61.10\% from the BoB RBF SVM model.
Results for all models are based on the same three fold cross-validation mean F1 as previously.

The FACS AUs SVM model performs better on my dataset than the DAIC dataset.
This is almost certainly due to the difference in quantity and quality of available FACS AU features.
The DAIC dataset provides 20 AU predictions per frame while my dataset provides 35 AU predictions.

\begin{table}[t]
  \begin{center}
    \begin{tabular}{rrrr}
      \toprule
      \multicolumn{2}{c}{\textbf{Model}} & \textbf{F1 avg} & \textbf{F1 std} \\ [0.5ex]

      \midrule
      \multirow{2}{*}{\texttt{lin}}
      & Optimal feature combination & \textbf{82.70\%} & 8.95\% \\
      & All features baseline & 66.81\% & 8.89\% \\
      \midrule
      \multirow{2}{*}{BoB}
      & RBF kernel & 61.10\% & 6.09\% \\ % \cmidrule{2-4}
      & Linear kernel & 60.13\% & 9.24\% \\
      \midrule
      \multirow{2}{*}{FACS AUs}
      & My dataset & 63.71\% & 8.02\% \\ % \cmidrule{2-4}
      & DAIC dataset & 56.53\% & \textbf{3.10\%} \\

      \bottomrule
    \end{tabular}
  \end{center}
  \caption{
  Comparison of my \texttt{lin} model, FACS AUs based SVM models, and the BoB SVM models for
  predicting \textbf{depression} as defined by the PHQ-8 questionnaire, on my dataset.
  The FACS AUs DAIC model is the exception, its results are for the DAIC dataset.
  The \texttt{lin} model using the optimal feature combination identified in
  Section~\ref{sec:results-feature-representation} achieves the best results.
  The \texttt{lin} model using all features (i.e.\ the all feature baseline from Section~\ref{sec:results-feature-representation})
  also beats the comparison methods.
  }
  \label{table:results-comparison-of-methods}
\end{table}

\section{Multi-Modal}\label{sec:evaluation-multi-modal}
Given the success of the FACS AUs SVM classifier in Section~\ref{sec:results-comparison-method},
I also evaluate a multi-modal method that fuses my gesture meta features and FACS AUs.

I perform multiple multi-modal experiments:
\begin{enumerate}
  \item Feature vector fusion including all features, evaluated across all four classifier types.
  \item Feature vector fusion with a feature search on my gesture meta features, though all AU features are
  retained in every search iteration.
  \begin{enumerate}
    \item Search using the \texttt{lin} classifier.
    \item Search using the \texttt{svm} classifier, as this was used successfully with the AUs features
    alone.
    \item Search using a radial-basis function kernel SVM classifier which achieves comparable
    results to the linear kernel SVM classifier on the AUs features alone.
  \end{enumerate}
  \item Hybrid fusion inspired by Alghowinem et al.\ \cite{Alghowinem:2015cs}.
  This involved feature fusion and decision level fusion via a majority vote of three classifiers:
  one classifier with only meta features, one with only AUs features, and one with the fused
  feature vector.
  I experimented with all meta features and the feature searched meta features.
  I used the best classifier type for the individual classifiers, i.e.\ \texttt{lin} for gesture meta features,
  \texttt{svm} for AUs features, and then also \texttt{svm} for fused features.
\end{enumerate}

\subsection*{Results}
The best result was by the feature search method with an \texttt{svm} classifier (2.b),
achieving 81.94\% F1 average.
The feature search with \texttt{lin} classifier (2.a) was second best with an 80.93\% F1 average.
However, these are worse than the best depression detection score of 82.70\% when using the gesture meta
features alone with the \texttt{lin} model.
Moreover, the hybrid fusion approaches achieved F1s in the mid-70s, so rather than correcting
errors by the meta features, it averaged the success rate between the meta classifier and
the AUs classifier, resulting in a worse F1.

\subsection*{Potential Future Fusion Approaches}
These approaches to fusion are somewhat simplistic.
More sophisticated approaches, such as deep learning fusion, may achieve better results than
the meta features alone.
An interesting route for future work is to use recurrent deep learning to fuse temporally aligned
meta features with AUs, rather than fusing them post-aggregation.

\section{Summary}\label{sec:evaluation-summary}
In this chapter I have evaluated the introduced gesture meta features and found they provide linear
predictive information.
To achieve the best performance I perform an exhaustive search of the feature space to identify the
best feature combination.
This demonstrates the potential of the features, however, it also introduces a risk of overfitting.
This risk is mitigated by two factors: the best performance is achieved by a linear regression
based classifier (i.e.\ a classifier not prone to overfitting) and many features are present in
optimal feature combinations for multiple uncorrelated labels.

I compared my method to a basic facial-modality method and an existing body-modality method based on
STIPs (i.e.\ generic analysis of video data).
My method outperforms both when using the \texttt{lin} classifier, both when using all possible features
and when using the optimal feature combination for the depression label.

Finally, I perform a multi-modal experiment utilising the facial-modality comparison method and my
novel gesture meta features.
Despite testing multiple approaches to fusion, no method I evaluated beat the mono-modal gesture
meta features classifiers.
However, all related work that I am aware of achieves better performance when incorporating multiple
modals, as such it is likely that further investigation of multi-modal approaches incorporating the
gesture meta features will identify a multi-modal approach that does improve upon my mono-modal results.

\chapter{Conclusion}\label{ch:conclusion}

I have presented a novel set of generic body modality features based on meta information of gestures.
To develop and evaluate these features I also introduced a novel non-clinical audio-visual dataset containing
recordings of semi-structured interviews along with distress and personality labels.

I evaluated these features for detecting depression as a binary classification task based on
PHQ-8 scores.
A linear regression based classifier achieved a 82.70\% average F1 score, suggesting these features
are both useful for depression detection and provide linear information.
I further evaluated the features' generalisability via binary classification tasks for other
distress labels, such as anxiety, personality measures, and gender.
All generalisation labels achieved better than 80\% average F1 scores, with the exception of one label,
personality neuroticism, which achieved 76.39\%.

These are novel features as previous works within the body modality for automatic distress detection
are based on either unsupervised definitions of expressions or hand-crafted descriptors of distress
indicative behaviour.
These features are similar to some of the work within the eye and head modalities which are based
on modeling generic activity levels within these modalities, among other features.

Finally, these features exist within a broader ecosystem of affect based features.
The best results for automatic distress detection, and similar tasks, is achieved by integrating
multiple modalities and multiple representations of modalities.
Future work may apply these features to new tasks, extend the feature type by extracting similar
meta features, and develop methods for combining them with other modalities to amplify the information
they provide.

\section*{Future Work}
\section*{Dataset}
Future work could expand the introduced dataset and increase its quality via manual annotations.
For example, annotating time frames based on the topic of conversation could enable sample segmentation
methods such as Gong \& Poellabauer~\cite{Gong:2017fj}.
This could improve the feature modeling and classifier stages, another route of future work is
gesture definition and detection.
For example, manual annotation of gestures would enable auto-learnt gesture detectors.

\section*{Regression Tasks}
All labels I evaluate, with the exception of gender, are scales based on self-evaluation questionnaires
relating to distress or personality.
The distress scales define multiple levels of severity while the personality scales do not.
Both types of labels are prime for regression prediction tasks.
Moreover, distress classification prediction could be provided by the defined severity levels once a
regression had been performed.
However, this requires a larger dataset (initial experiments with regression on the dataset support
this assertion).

\section*{Features}
Applying the presented features within larger datasets could further illuminate the properties of certain
features, either confirming their inconsistency, or solidifying the linearity in one direction.
Such properties should then be tested with regards to the psychology literature.

\subsection*{Trunk}
I do not evaluate a \enquote{trunk} localisation (i.e.\ hips to shoulders), though it may prove
useful in future work.
In my dataset participants are seated during the whole interview, thus the two-dimensional range of
motion in the trunk (i.e.\ up/down and side to side) is small.
As I do not include depth recording the greatest range of motion, forward and backward, is difficult
to incorporate.
Future work that either includes depth sensing or has participants in a wider variety of scenarios,
where they may be standing or more physically engaged, may find some use in a trunk localisation.

\subsection*{Cross Localisation Co-occurrence}
Co-occurrence of gestures across localisations is not considered in this dissertation, though it is
an area for future work.
Applying deep learning to co-occurrence modeling to identify relevant patterns would be particularly
interesting.

\subsection*{Improved Surprise}
In future work this feature could be extended to be more indicative of \enquote{surprise}, rather
than simply an average of distance since last gesture.
For example, a gesture that occurs within a regular pattern of gestures, regardless of their distance,
might be considered to have low surprise, while a gesture interrupting that pattern by occurring in the
middle of the pattern's silent period would be very surprising.
This more sophisticated measure of surprise could account for naturalistic behaviour such as
rhythmic movement/gestures.
These types of extensions of features are exciting for two reasons:
firstly, they are interpretable, they do not come from a black box model, and thus can be supported
by psychology literature.
Secondly, they are immediately measurable and their design is based on traditional statistical techniques
such as repetition modeling and can therefore be iterated on more readily than deep learning based
features.

\subsection*{Generalisation}
Future work, some of which has been discussed here, could extend these features to achieve better
results in more diverse datasets.
A greater variety of scenarios where the participant is more constantly physically engaged
such that movement is more constant, e.g\ walking or a standing conversation, would challenge
the design of the presented gesture meta features.
Indeed, the current design would have trouble as it relies on the default state being a lack of movement.

\subsection*{Deep Learning for Feature Generation}
As I discussed in Chapter~\ref{ch:literature-review}, deep learning has been applied to the depression
detection task, but still relies on hand-crafted features and descriptors.
As with many fields it is likely that deep learning methods will, in the next few years, achieve
state-of-the-art results, far exceeding the potential of more traditional approaches.
However, to reach this point the deep learning models need large enough datasets to train on.

Assuming no especially large dataset is developed specifically for automatic depression detection, one potential
for future work is the use of transfer learning (as Chen et al.\ \cite{Chen:2017ca} explore for
vocal and facial features) for body expressions.
For example, CNNs could be trained on general body expression datasets to learn inner representations
of body language.
The output of inner layers could then be used in depression detection tasks
(Razavian et al.\ \cite{Razavian:wa} present a transfer learning approach to image recognition tasks
and achieve very good results on niche tasks using a generic pre-trained image CNN).

\section*{Classifiers and Deep Learning}

In this dissertation I have used two traditional statistical models, linear and logistic regression,
and two more advanced machine learning methods, support vector machines and random forests.
However, I have purposefully avoided more complex machine learning methods such as neural networks,
and especially deep learning.
These more sophisticated methods have achieved significant improvements on the state-of-the-art in
many domains, however, they suffer from a lack of interpretability and a propensity to overfit data.
As my primary focus has been validating gesture meta features I define interpretability as a core
requirement and, given the size of my dataset, avoiding methods that overfit is important.

However, having validated these features, deep learning presents opportunities with regards to learning
more sophisticated aggregation functions (this is slightly different to the feature generating deep
learning discussed above).
I aggregate three core features of gestures: their movement, duration, and \enquote{surprise}.
I perform aggregation via averaging and standard deviations.
However, given a larger dataset, a deep learning model could foreseeably learn both more sophisticated
aggregation functions and core meta-features.
It is important to emphasize here the burden on dataset size that exists when attempting to learn
valuable features using deep learning.

\subsection*{Multi-Modal Approaches with Deep Learning}
Deep learning can learn useful modality integration functions.
While I experimented with a multi-modal approach in Section~\ref{sec:evaluation-multi-modal} it did
not achieve better results than my mono-modal method.
I believe this is due to the rather direct approaches to multi-modal integration I experimented with.
Applying deep learning to modal integration thus presents an opportunity for further work.

Furthermore, integrating a greater diversity of modalities and modality representations would be
interesting.
For example, applying the Bag-of-Body Dynamics features along with my gesture meta features and FACS
AUs.
With regards to the Bag-of-Body Dynamics method, future work could apply the core clustering concept to pose
estimation data rather than STIPs data and may achieve better results from this.

  \cleardoublepage
  \phantomsection
  \addcontentsline{toc}{chapter}{Bibliography}
  \bibliographystyle{plainnat}
  \bibliography{papers_full,dissertation_misc}

  \printthesisindex

\end{document}